\documentclass{article}

\usepackage[preprint,nonatbib]{neurips_2026}

\usepackage[utf8]{inputenc} 
\usepackage[T1]{fontenc}    
\usepackage{hyperref}       
\usepackage{url}            
\usepackage{booktabs}       
\usepackage{amsfonts}       
\usepackage{nicefrac}       
\usepackage{microtype}      
\usepackage{xcolor}         
\usepackage{graphicx}
\usepackage{caption}
\usepackage{tabularx}
\usepackage{listings}
\usepackage{soul}
\usepackage{capt-of}
\usepackage[shortlabels]{enumitem}
\usepackage{dsfont}
\usepackage{multirow}
\usepackage{adjustbox}
\usepackage{arydshln}
\usepackage{algorithm}
\usepackage{algpseudocode}
\usepackage{amsmath}
\usepackage{amssymb}
\usepackage{mathtools}

\usepackage{graphicx}
\usepackage{subcaption}
\usepackage{float}
\usepackage{makecell}
\usepackage{tikz}
\usepackage{wrapfig}
\usetikzlibrary{positioning, arrows.meta, trees}

\hypersetup{hidelinks}

\tikzset{
  kdnode/.style={draw, rounded corners, align=center, font=\small, minimum width=2.2cm, minimum height=0.8cm},
  cond/.style={kdnode, fill=gray!10},
  leaf/.style={kdnode, fill=blue!5},
  edge/.style={-Latex},
  edge from parent/.style={draw, -Latex},
  level distance=1.6cm,
  level 1/.style={sibling distance=5cm},
  level 2/.style={sibling distance=3.5cm},
  level 3/.style={sibling distance=2.5cm}
}

\usepackage{xcolor}

\renewcommand\arraystretch{1.2}

\title{CDRL: Certification-Driven Reinforcement Learning for Neutrino Flavor Model Discovery}

\author{%
  Piyush Jha$^{1,2}$ \quad Jake Rudolph$^{3}$ \quad Victoria Knapp-P\'erez$^{3,6}$ \quad
  Max Fieg$^{4}$ \\
  \textbf{Aishik Ghosh}$^{2,5}$\textbf \quad \textbf{Vijay Ganesh}$^{1}$ \\[4pt]
  $^{1}$School of Computer Science, Georgia Institute of Technology, USA \\
  $^{2}$School of Physics, Georgia Institute of Technology, USA \\
  $^{3}$University of California, Irvine, USA \quad
  $^{4}$Particle Theory Department, Fermilab, USA \\
  $^{5}$Lawrence Berkeley National Laboratory, USA \\
  $^{6}$Halluminate, Department of Research, USA \\[2pt]
\texttt{piyush.jha@gatech.edu, jirudolp@uci.edu, vknapppe@uci.edu},\;\\
\texttt{mfieg@fnal.gov, aishikghosh@physics.gatech.edu, vganesh@gatech.edu}
}

\begin{document}

\maketitle

\begin{abstract}
Many scientific discovery problems require searching combinatorial hypothesis spaces under complex domain constraints. Reinforcement learning (RL) offers a promising approach, but existing methods rely on scalar rewards that provide limited information about why candidate solutions fail, leading agents to repeatedly explore invalid regions.
We introduce Certification-Driven Reinforcement Learning (CDRL), a framework that leverages structured feedback from symbolic reasoning tools. When a candidate violates domain constraints, these tools produce certificates identifying the {\it actions} responsible for failure. CDRL converts these certificates into reusable constraints that eliminate classes of invalid solutions and guide exploration toward valid regions.
We evaluate CDRL on neutrino flavor model discovery in theoretical particle physics, where the hypothesis space exceeds $10^{26}$ possible models, and compare it with the state-of-the-art RL approach previously used for this task. Across three theory spaces, CDRL achieves up to 1.95$\times$ higher valid model rates and up to 6.33$\times$ higher neutrino model rates while evaluating up to 4$\times$ fewer candidates. We further extract 40 interpretable rules from search trajectories using a post-hoc decision-tree framework and show that reusing them as soft constraints yields gains of up to 2$\times$ in valid model rates and 3$\times$ in neutrino model discovery across all three theory spaces. These results suggest that CDRL uncovers reusable structure in combinatorial search spaces and provides a general framework for scientific model discovery.
\end{abstract}

\section{Introduction}

Scientific discovery often requires exploring large spaces of candidate hypotheses. Researchers propose mathematical models, derive predictions, and compare them with experimental observations. This iterative process lies at the core of many scientific disciplines, including particle physics~\cite{baretz2025towards,feruglio2021leptonflavoursymmetries,Almumin:2022rml,degouvea2022theoryneutrinophysics}, computational biology~\cite{jumper2021highly}, chemistry~\cite{segler2018planning}, and materials science~\cite{bai2025artificial}. In many such settings, the space of candidate models is combinatorial, and small changes in model structure can lead to dramatically different predictions. Systematically exploring these large theoretical spaces therefore becomes a major computational challenge, motivating the development of AI systems that assist scientific discovery and exploration~\cite{merchant2023scaling,abramson2024accurate,mankowitz2023faster}.

A common approach to exploring large hypothesis spaces is reinforcement learning (RL)~\cite{sutton_rl}, which formulates discovery as a sequential decision-making problem. In these frameworks, an agent incrementally constructs candidate models and receives rewards based on whether they satisfy domain-specific criteria. RL has been applied successfully to automated theorem proving~\cite{li2024survey}, program synthesis~\cite{zhang2025advances}, and scientific discovery~\cite{wang2023scientific,martin2021reinforcement,gow2022review}. However, these methods typically rely on scalar reward signals that provide little information about why a candidate fails~\cite{jha2024rlsf}. Consequently, agents often waste substantial effort repeatedly exploring similar invalid regions of the combinatorial search space.

\begin{figure}[t]
    \centering
    \includegraphics[width=\textwidth]{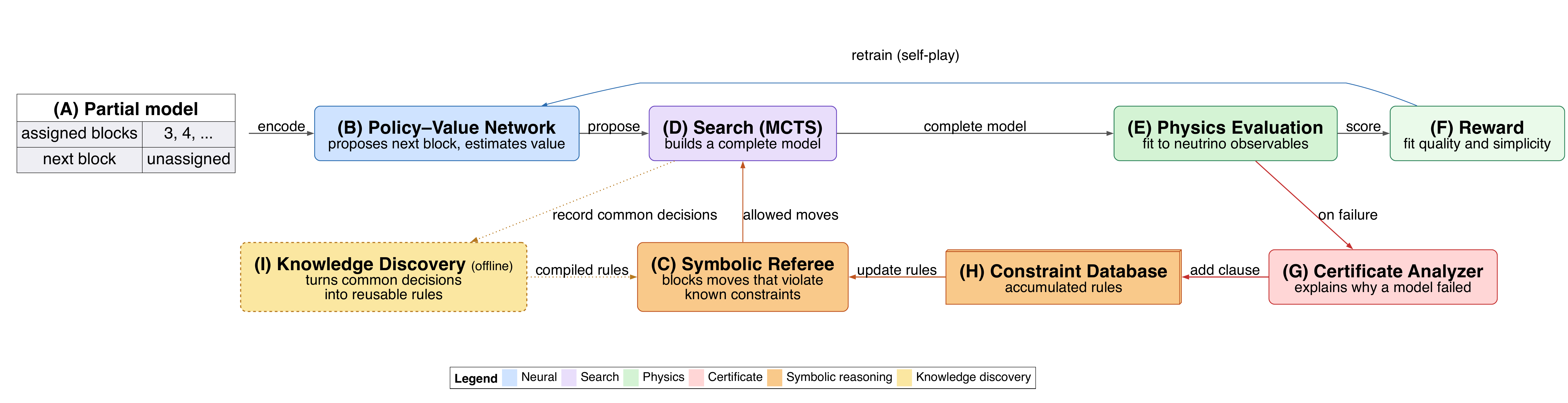}
    \caption{\textbf{The CDRL loop.} CDRL builds each model one block at a time (A) using a policy-value network and search (B, D)~\cite{silver2016mastering,Silver2017}. Before the search commits to a block, a symbolic referee (C) rules out moves that are already known to violate a domain constraint, so search time is spent only on moves that remain possible. Once a model is complete, it is scored by the physics pipeline (E, F; Section~\ref{subsec:env_and_rew} defines the score and what counts as a good model). When a model fails, the certificate analyzer (G) identifies which specific block assignments caused the failure and compiles that into a reusable rule stored in the constraint database (H), so the referee (C) blocks the same failure pattern, and every equivalent model that would have made it, everywhere it recurs in future search. Separately, and only after training, search knowledge discovery (I) mines the most common decisions the search made into simple reusable rules that also feed the referee (C). Section~\ref{sec:sat} gives the formal, SAT-based definition of the referee and constraint database, and Table~\ref{tab:cert_types} works out the ways a certificate can arise. Figure~\ref{fig:mechanism_tree_steps} (Appendix~\ref{app:mechanism_steps}) walks through this loop step by step on a worked example.}
    \label{fig:cdrl_arch}
\end{figure}

\begin{figure}[t]
    \centering
    \includegraphics[width=0.7\textwidth]{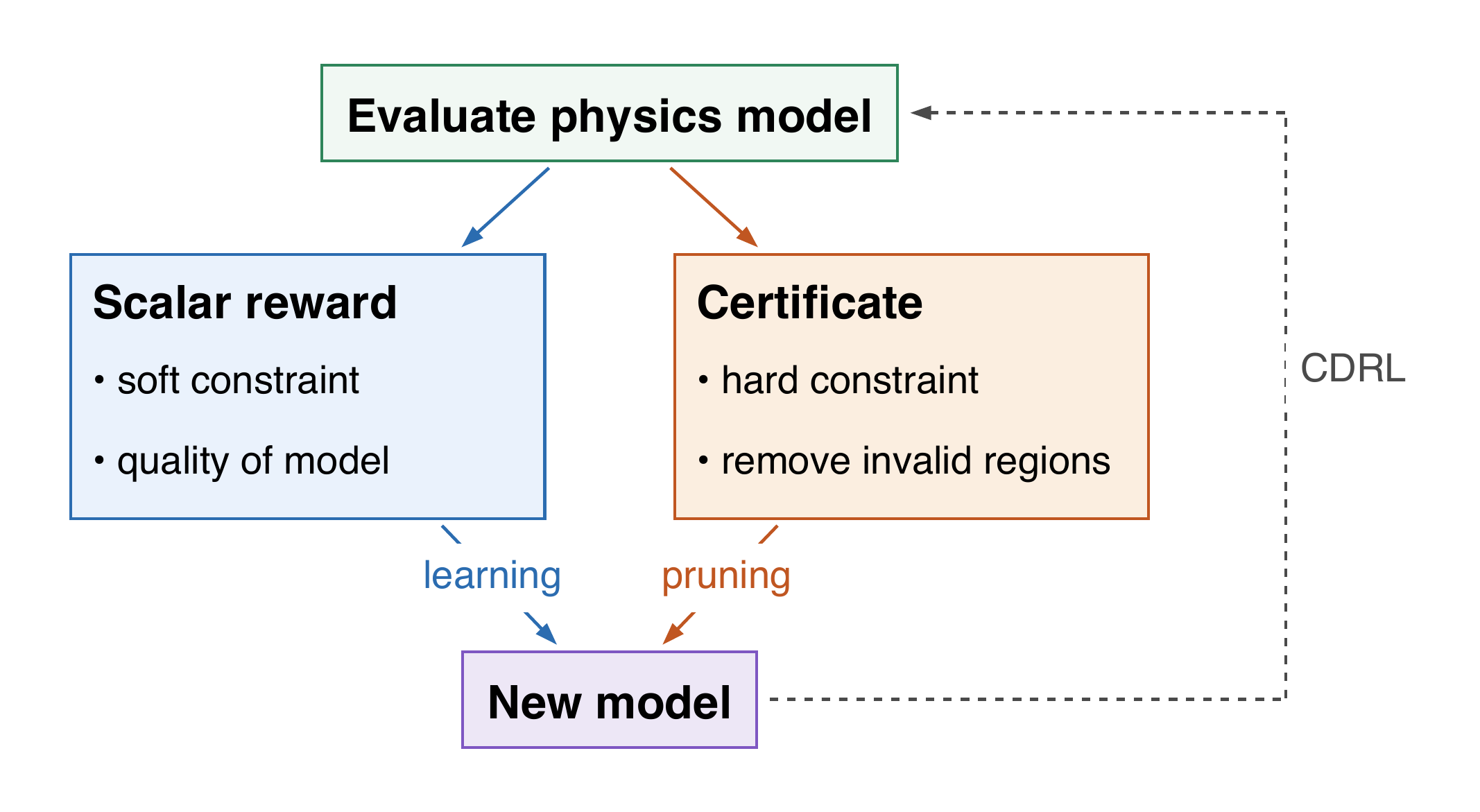}
    \caption{\textbf{The two feedback signals CDRL gets from each physics evaluation are complementary rather than competing.} Evaluating a candidate model yields a \emph{scalar reward}, a \emph{soft} constraint that scores the quality of the model, and a \emph{certificate}, a \emph{hard} constraint that names the exact components responsible for a failure and forbids them via a clause that Boolean Constraint Propagation (BCP) then enforces everywhere that pattern would recur. The reward shapes the network through \emph{learning}, while the certificate \emph{prunes} the search space; together they determine the new model the search proposes next, and the cycle repeats. Section~\ref{sec:cdrl} gives the formal definition of a certificate.}
    \label{fig:feedback_complementarity}
\end{figure}

In many scientific domains, structured feedback beyond scalar rewards is available, yet traditional RL algorithms do not fully exploit it. External reasoning tools such as theorem provers, constraint solvers, computer algebra systems, program analysis tools, and simulators can analyze candidate solutions and identify the causes of failure. Because these systems operate on formally defined objects, they can often produce \textit{certificates}~\footnote{A certificate is a structured explanation produced by a reasoning tool that identifies the subset of decisions responsible for satisfying or violating domain constraints. See Section~\ref{sec:cdrl} for a formal definition and discussion.} that describe the outcome of their analysis and pinpoint the components responsible for violated constraints. Such certificates provide structured information that can be leveraged to guide exploration more effectively than scalar reward signals alone. Crucially, the two signals are complementary rather than competing (Figure~\ref{fig:feedback_complementarity}). 

Further, scientists have traditionally been able to distill useful heuristic rules from the problem-solving process, and these insights can accelerate future exploration of the underlying mathematical space. An RL approach that enables the extraction of such heuristic rules from the agent's decisions would therefore be valuable, both for interpretability and for improving computational efficiency in subsequent runs.

There is a growing interest in using RL to construct theories in particle physics~\cite{Wojcik:2024lfy,baretz2025towards,Alexander:2026lpw,Giarnetti:2025mit}. A viable theory, or model, is subject to a number of consistency checks such as requiring that Lorentz invariance and other internal symmetries are respected, as well as demands that the theory reproduce certain observed qualities like the mass of a certain particle. When an RL agent attempts to construct a model, it will generically fail these checks. However, if the RL agent had feedback in the form of a certificate concerning the failure of a candidate model, it could be trained to build viable models more quickly. To demonstrate the potential advantages of fully incorporating such feedback, we consider models that can be constructed for neutrino masses~\cite{baretz2025towards}, which enables rapid exploration of mathematical models where physicists have limited intuition.

In our setup, an agent selects the particle content of a model and the symmetries it must satisfy. These constraints are encoded via irreducible representations of a symmetry group, yielding a Lagrangian density that defines a quantum field theory. The resulting models are evaluated using existing symbolic and numerical software for Lagrangian construction, mass-matrix extraction, and parameter fitting, followed by statistical comparison to experimental data. Exhaustive search over more than $10^{26}$ possible models is infeasible, as evaluating a single candidate requires non-trivial symbolic and numerical computation.

Certification-Driven Reinforcement Learning (CDRL) is a framework that leverages structured feedback from external reasoning tools to improve exploration in constrained combinatorial search spaces. When a candidate violates a domain constraint, the reasoning tool returns a certificate identifying the subset of assignments responsible for the failure. CDRL analyzes this certificate and converts the responsible assignments into a symbolic conflict clause, which is added to the global SAT constraint database~\cite{biere2009handbook}. During search, these learned constraints are enforced by a lightweight symbolic reasoning layer that immediately eliminates partial candidates violating known domain rules, allowing the RL agent to avoid exploring impossible regions of the search space. 

During the MCTS rollouts that follow~\cite{silver2016mastering,Silver2017}, this symbolic reasoning layer is implemented via Boolean Constraint Propagation (BCP)~\cite{davis1962machine,biere2009handbook}, which enforces the accumulated clauses as the agent builds new candidates, blocking exploration of any partial assignment that repeats a previously seen conflicting decision pattern. This differs from a negative reward in \emph{kind}, not degree. A negative reward is a scalar that adjusts the network's weights by a small gradient step; it lowers the \emph{probability} of a decision but never rules it out, so the agent can, and empirically does, revisit the same dead end many times, and the penalty says nothing about \emph{which} of the decisions was to blame. A conflict clause instead operates on the search space itself: it is a hard logical constraint naming the exact assignments responsible, and once added, BCP makes every partial assignment containing that pattern \emph{unreachable} for all agents sharing the clause database, with no dependence on how well the network has been trained. In effect, a scalar reward changes where the agent \emph{prefers} to go, whereas a certificate changes where it \emph{can} go. As more certificates accumulate, the feasible search space is progressively pruned, allowing the agent to avoid rediscovering equivalent failure modes and focus expensive physics evaluations on regions more likely to contain valid theories.

As a concrete example of this structured feedback in our setting, consider a candidate neutrino model whose non-Abelian $A_4$ representation assignment is invalid for \emph{any} choice of $\mathbb{Z}_N$ charge (Figure~\ref{fig:cdrl_arch}). Rather than emitting only a scalar penalty for this one candidate, our system extracts a \emph{certificate} identifying the representation pattern responsible and compiles it into a logical constraint that blocks that pattern across the whole charge sector, so it is masked the moment it reappears elsewhere in the search tree, before any physics evaluation is run on it (worked out in detail in Section~\ref{sec:application}). Concretely, we realize this loop with well-established tools: candidate models are encoded as a Boolean satisfiability (SAT) formula in conjunctive normal form; certificates become \emph{conflict clauses} that are enforced during search by Boolean Constraint Propagation (BCP)~\cite{davis1962machine,biere2009handbook}, the same unit-propagation mechanism used inside modern SAT solvers~\cite{biere2009handbook}; and the search itself is an AlphaZero-style Monte Carlo Tree Search (MCTS)~\cite{silver2016mastering,Silver2017} guided by a policy-value network. These three components play distinct, complementary roles at every decision: neural guidance \emph{proposes} the next assignment, symbolic reasoning \emph{prunes} the assignments a certificate has ruled out, and MCTS \emph{balances} exploration against exploitation over what remains. We quantify the payoff of this structured feedback empirically across three neutrino theory spaces (Section~\ref{sec:results}), and further show that the constraints and decision patterns CDRL accumulates expose reusable structure in the theory space (Section~\ref{sec:kd_results}).

Across three theory spaces for neutrino flavor model discovery, CDRL achieves up to $1.95\times$ higher valid model rates and up to $6.33\times$ higher neutrino model rates than AMBer~\cite{baretz2025towards}, while evaluating up to $4\times$ fewer candidates. These gains are expected to scale with the complexity of the problem and indicate substantially more efficient navigation of the underlying combinatorial hypothesis space, similar in spirit to recent advances in algorithm discovery and search-based scientific reasoning~\cite{fawzi2022discovering,ruiz2025quantum,mankowitz2023faster,abramson2024accurate}.

Beyond improved search efficiency, CDRL enables search knowledge discovery from search trajectories. Using a post-hoc decision-tree analysis of high-confidence MCTS states, we extract 40 interpretable rules that capture recurring dependencies between particle assignments. Reusing these rules as soft constraints further improves performance, suggesting that they capture meaningful and reusable structure within the theory space.

\textbf{Our contributions} are therefore: (i) CDRL, a certificate-driven RL framework that transforms symbolic failure certificates into reusable search constraints, progressively eliminating large invalid regions; (ii) state-of-the-art results on neutrino flavor model discovery, achieving substantially higher discovery rates with fewer evaluations than prior RL approaches; and (iii) a framework for extracting and reusing interpretable rules from search trajectories, demonstrating that knowledge acquired during exploration can improve future search.

\begin{table}[t]
\centering
\caption{\textbf{Example neutrino flavor model discovered by CDRL in the $T_{19} \times \mathbb{Z}_4$ theory space.} The table shows particle representation and charge assignments for one of the viable neutrino models discovered by CDRL. This model achieves a strong $\chi^2$ fit with a minimal number of free parameters, illustrating the ability of CDRL to discover compact and high-quality theories. Additional discovered neutrino models across all three theory spaces are provided in Section~\ref{app:kd_rules}.}
\label{tab:t19z4_model}
\setlength{\tabcolsep}{5pt}
\renewcommand{\arraystretch}{1}
\scriptsize
\begin{tabular}{c|ccccccccccccc}
\toprule
 & $L$ & $E_1$ & $E_2$ & $E_3$ & $N$ & $H_u$ & $H_d$
 & $\phi_1$ & $\phi_2$ & $\phi_3$ & $\phi_4$ & $\phi_5$ & $\phi_6$ \\
\midrule
$T_{19}$
& $\mathbf{3}_1$
& $\mathbf{1}''$
& $\mathbf{1}$
& $\mathbf{1}'$
& $\bar{\mathbf{3}}_2$
& $\mathbf{1}'$
& $\mathbf{1}$
& $\mathbf{3}_1$
& $\bar{\mathbf{3}}_1$
& $\mathbf{3}_1$
& $\bar{\mathbf{3}}_1$
& $\mathbf{3}_2$
& $\mathbf{3}_2$ \\

$\mathbb{Z}_4$
& $4$
& $2$
& $2$
& $2$
& $4$
& $2$
& $4$
& $2$
& $2$
& $3$
& $4$
& $4$
& $1$ \\
\bottomrule
\end{tabular}
\end{table}

\section{Background and Related Work}\label{sec:background}

\noindent{\bf Neutrino Flavor Model Construction.}
The Standard Model (SM) of particle physics describes the fundamental particles and their interactions and has exquisite agreement with experiment, with some exceptions. Perhaps the most notable exception is the prediction of massless neutrinos, but experiments~\cite{Super-Kamiokande:1998kpq,SNO:2002tuh,Cleveland:1998nv, K2K:2002icj} have determined that they have a mass, which necessitates the presence of new particles beyond the SM. Because these new particles must couple to neutrinos, and because neutrinos and charged leptons are components of the same electroweak doublets, one generally expects them to couple to charged leptons as well. For this reason, solutions to the massive neutrino problem are typically connected to the question of flavor in the SM~\cite{Babu:2009fd,feruglio2021leptonflavoursymmetries,Altmannshofer:2024hmr,King:2015aea}: what is the explanation for the three flavors of fermions and what is the origin for the observed levels of mixing between them?

One general framework to address these shortcomings is to add right-handed neutrinos to generate small neutrino masses via the seesaw mechanism~\cite{Minkowski:1977sc,Yanagida:1979as,Glashow:1979nm,Gell-Mann:1979vob}, as well as a number of scalar particles, ``flavons'', whose interactions give structure to the level of lepton mixing after they acquire a certain vacuum expectation value (VEV). Adding these new particles generally results in an even larger number of free parameters than was already present in the SM, which limits the predictivity of the model. However, the parameters in the flavor sector can be strongly constrained if the model is subject to a global flavor symmetry, which gives structure to the observed mixing; in the ideal case it explains the observed values with minimal free parameters. A given theory under this framework is thus described by: the set of particles, the irreducible representations they take under the flavor symmetry, the VEV of the flavons\footnote{In a complete model, the VEV is a dynamical result of the theory, and achieving a certain VEV pattern is a problem in its own right. Here, as in Ref~\cite{baretz2025towards}, we do not attempt to solve this problem, but assume that a certain set of VEVs can be achieved}, and the remaining free parameters. The number of models that exist for a given choice of flavor symmetry grows rapidly with the number of particles, and identifying solutions that fit to experiment is highly nontrivial.

Recent work~\cite{Halverson_2019,Harvey:2021oue,Nishimura:2020nre,Nishimura:2024apb,Carta:2025asr} has demonstrated RL as a tool for searching analogous spaces of models for various areas in particle physics. In particular, the AMBer framework~\cite{baretz2025towards} integrates domain specific physics tools which allows the agent to evaluate candidate models using software used by physicists for conventional model-building.
CDRL goes further: rather than using these tools only to evaluate candidates one at a time, it extracts structured certificates from them and converts these into reusable constraints that prune invalid regions of the search space, letting the system progressively accumulate knowledge about the structure of valid theories. This approach can be extended to several other applications of agent-guided workflows to scientific discovery.

\noindent{\bf AI for scientific discovery.}
Artificial intelligence is increasingly used across the scientific discovery pipeline, including hypothesis generation, experimental design, and large-scale data analysis~\cite{wang2023scientific,xu2021artificial}. Recent advances highlight the ability of AI systems to explore large scientific search spaces, such as materials discovery using graph neural networks and generative models~\cite{merchant2023scaling,liu2023generative}, and biomolecular structure prediction using deep learning systems like AlphaFold~\cite{abramson2024accurate}. Machine learning is also widely used in particle physics for analyzing experimental data and identifying potential new phenomena~\cite{karagiorgi2022machine}. Alongside these developments, there is growing interest in incorporating domain knowledge and reasoning into AI-driven discovery processes~\cite{messeri2024artificial,van2023ai}. In this work, we take a step in this direction by introducing CDRL, which integrates symbolic reasoning tools into the learning loop to provide structured and interpretable feedback during exploration.

\noindent{\bf Reinforcement learning for algorithm and scientific discovery.}
Reinforcement learning (RL) has discovered algorithms and optimized complex scientific systems, including AlphaTensor for tensor decomposition~\cite{fawzi2022discovering}, AlphaTensor-Quantum for fault-tolerant quantum circuits~\cite{ruiz2025quantum}, and AlphaDev for faster sorting~\cite{mankowitz2023faster}. It has also controlled physical systems and experiments such as tokamak plasmas~\cite{degrave2022magnetic}, stratospheric balloons~\cite{bellemare2020autonomous}, and quantum experiment design~\cite{melnikov2018active}. These results show RL can navigate large combinatorial search spaces, but they depend on reward signals from a simulator or environment. CDRL instead adds symbolic reasoning tools that emit structured certificates, giving the agent richer feedback to guide exploration.

\noindent{\bf Neuro-symbolic AI.}
Neuro-symbolic AI combines neural learning with symbolic reasoning to strengthen reasoning, interpretability, and reliability~\cite{hitzler2023compendium,acharya2023neurosymbolic}. Integration strategies range from embedding symbolic rules within neural architectures and differentiable reasoning modules to modular systems that call external reasoning tools~\cite{colelough2025neuro}. CDRL takes the modular route: a learning agent queries external symbolic systems that analyze candidate solutions and return structured feedback. In the taxonomy of~\cite{colelough2025neuro}, CDRL sits at the center of the neuro-symbolic design space, uniting learning, symbolic reasoning, and structured knowledge within a single decision loop.

\noindent{\bf Oracle-augmented learning.}
Recent work has explored integrating external reasoning tools into AI systems. Large language models have been combined with symbolic tools during inference~\cite{kambhampati2024llms,agrawal2024monitor,chen2020program,jana2023attention}, while other approaches incorporate external feedback during RL~\cite{guo2025deepseek,jha2024rlsf}. 
Related ideas also appear in automated reasoning and synthesis. Modern SAT solvers use conflict feedback to guide search~\cite{liang2016learning}, and frameworks such as CEGIS and Satisfiability Modulo Oracles refine candidates using counterexamples or constraints from external verifiers~\cite{jha2010oracle,polgreen2022satisfiability,jha2017theory}. 
CDRL builds on these directions by converting certificates from reasoning systems into reusable constraints, enabling RL agents to progressively prune invalid regions of large scientific search spaces.

Together, these observations motivate Certification-Driven Reinforcement Learning (CDRL), a paradigm in which symbolic reasoning systems act as structured feedback generators that guide RL over large scientific search spaces.

\section{Certification-Driven Reinforcement Learning (CDRL)}
\label{sec:cdrl}

\begin{algorithm}[t]
\caption{Certification-Driven Reinforcement Learning (CDRL)}
\label{alg:cdrl}
\begin{algorithmic}[1]

\Statex \textbf{Notation:} $\pi_\theta$ policy-value network; $V$ verifier (the physics evaluation pipeline); $\mathcal{A}$ certificate analyzer; $\mathcal{K}_0, \mathcal{K}$ initial and accumulated constraint sets; $H$ space of candidate hypotheses (models); $h \in H$ one candidate; $\mathcal{H}_{\text{valid}}$ set of discovered valid hypotheses; $\sigma$ a certificate returned by $V$ on failure; $c$ the constraint derived from $\sigma$.
\Statex

\Require Policy $\pi_\theta$, verifier $V$, certificate analyzer $\mathcal{A}$, initial constraints $\mathcal{K}_0$
\Ensure Set of valid theories $\mathcal{H}_{\text{valid}}$

\State $\mathcal{K} \leftarrow \mathcal{K}_0$ \Comment{initialize with known domain constraints}
\State $\mathcal{H}_{\text{valid}} \leftarrow \emptyset$ \Comment{store discovered valid solutions}

\For{each training episode}

    \State $h \sim \textsc{Plan}(\pi_\theta, \mathcal{K})$
    \Comment{construct candidate while respecting accumulated constraints}

    \State $(status, \sigma) \leftarrow V(h)$
    \Comment{evaluate candidate and optionally return failure certificate $\sigma$}

    \If{$status = \text{valid}$}
        \State $\mathcal{H}_{\text{valid}} \leftarrow \mathcal{H}_{\text{valid}} \cup \{h\}$
        \Comment{record valid hypothesis}
        \State Store trajectory and reward in replay buffer
    \Else
        \State $c \leftarrow \mathcal{A}(\sigma)$
        \Comment{convert certificate into reusable constraint}
        \State $\mathcal{K} \leftarrow \mathcal{K} \cup \{c\}$
        \Comment{prune entire class of invalid solutions}
        \State Store trajectory and penalty in replay buffer
    \EndIf
    
    \State Periodically update $\pi_\theta$ from replay-buffer mini-batches

\EndFor

\end{algorithmic}
\end{algorithm}

Many scientific discovery tasks can be viewed as structured search problems over a large space of candidate hypotheses. An agent incrementally constructs candidate solutions while interacting with an environment that verifies whether the candidate satisfies domain constraints.

\textbf{Certificate-Driven Reinforcement Learning (CDRL)} augments this RL loop with certificate-based constraint learning. Each time the verifier rejects a candidate, the certificate it returns is also converted into a reusable constraint that rules out the whole class of similarly invalid candidates, not just the one just tried.

Formally, let $H$ denote the space of candidate hypotheses and let $\mathcal{K}_0$ denote a set of initial domain constraints known a priori. The agent maintains a growing set of constraints $\mathcal{K}$ initialized with $\mathcal{K}_0$. During training, the agent proposes candidate hypotheses using a planning procedure $\textsc{Plan}(\pi_\theta, \mathcal{K})$ (such as MCTS), which incrementally constructs candidates while enforcing constraints $\mathcal{K}$ and using $\pi_\theta$ to guide exploration. Each candidate is verified by an external reasoning tool. If the candidate satisfies all constraints, the agent receives a positive reward. Otherwise, the verifier produces a certificate describing the failure. This certificate is analyzed to derive a new constraint, which is added to $\mathcal{K}$ and used to prune the search space.

The overall CDRL procedure is shown in Algorithm~\ref{alg:cdrl}. As training progresses, the learned constraints accumulate and progressively restrict exploration to regions of the hypothesis space that remain consistent with previously discovered knowledge. This allows the RL agent to avoid rediscovering similar failure modes and focus exploration on configurations that are more likely to yield valid solutions.

\noindent{\bf Certificates and Constraint Learning.}
A certificate is a structured explanation produced by a reasoning tool when a candidate hypothesis violates domain constraints. It identifies the subset of decisions responsible for the failure.
Let $h \in H$ denote a candidate hypothesis. When $h$ is invalid or redundant with an already-covered equivalent hypothesis, the verifier returns a certificate $\sigma$, which is analyzed to derive a constraint $c$ that prevents similar configurations. This constraint is added to $\mathcal{K}$ and enforced during future exploration, progressively pruning the search space.
This mechanism is conceptually related to conflict-driven learning in SAT solvers, where conflicts yield clauses that eliminate inconsistent assignments. In CDRL, certificates play an analogous role by enabling reusable constraint learning.

\section{Application: Neutrino Flavor Model Discovery}
\label{sec:application}
We evaluate CDRL on the discovery of viable neutrino flavor models, a central problem in particle physics. The task involves constructing models that explain neutrino masses and mixing while satisfying domain constraints. We follow a common flavor symmetry model building framework, where candidate models are defined by non-Abelian symmetry choices, particle irreducible representation assignments, Abelian charges, and vacuum expectation values. This results in a large combinatorial search space, making efficient exploration critical. Figure~\ref{fig:cdrl_arch} shows how the components below fit together; further background is provided in Section~\ref{sec:background}.

\subsection{Matrix Representation of Candidate Models}

Following the AMBer framework~\cite{baretz2025towards}, we encode a candidate theory as a matrix representation that captures the assignment of particles to symmetry representations and charges. Rows correspond to particles in the model, while columns correspond to irreducible representation and charge assignments under the symmetry groups. For example, rows include the three charged lepton doublets $(L_1,L_2,L_3)$, charged lepton singlets $(E_1,E_2,E_3)$, right-handed neutrinos $(N_1,N_2,N_3)$, Higgs fields $(H_u,H_d)$, and several flavon fields. Columns represent discrete non-Abelian representations (e.g., $A_4$ irreducible representations), Abelian $\mathbb{Z}_N$ charges, and VEV configurations for flavon fields. Columns are encoded using one-hot vectors. Unassigned entries are represented by $-1$, allowing the state to represent partially constructed models. This matrix therefore forms the state representation for the learning agent.

\subsection{Sequential Construction of Models}

Model construction is formulated as a sequential decision process. Instead of assigning every matrix entry independently, decisions are organized into blocks corresponding to logical modeling choices. The agent starts by assigning the active $\mathbb{Z}_N$ symmetry as one block, then for each particle in turn assigns its representation and then its Abelian charge as two separate blocks, and finally assigns each flavon's VEV configuration as its own block. At each step, the next unassigned block is selected, and the agent chooses from a small alphabet consisting of either an all-zero vector (inactive block) or a one-hot vector indicating the assignment. This structured action space reduces the branching factor and enables incremental construction of candidate theories until completion.

\subsection{SAT-Based Constraint Layer}
\label{sec:sat}

Rather than allowing the RL agent to freely construct candidate models, discovering only at the end that they violate domain rules, we incorporate a symbolic constraint layer that continuously enforces known scientific constraints during search. This layer immediately rejects partial assignments that cannot be extended into a valid theory, allowing the agent to focus its exploration on feasible regions of the combinatorial search space.

We encode the domain knowledge as a set of Boolean constraints, including exactly one representation per particle, a single active $\mathbb{Z}_N$ symmetry (a modeling choice that reduces the search space), exactly one charge per particle for that active $\mathbb{Z}_N$ symmetry, structural constraints for triplet representations, and exactly one vacuum expectation value (VEV) assignment per flavon. These constraints are compiled once into conjunctive normal form (CNF).

During the MCTS search, we use Boolean Constraint Propagation (BCP)~\cite{davis1962machine,biere2009handbook} at every node to maintain consistency between the current partial assignment and the encoded constraints. Whenever the agent makes a new assignment, BCP efficiently detects whether the partial model already violates a constraint and automatically infers any assignments that are logically forced. As a result, invalid branches are pruned immediately, while the state presented to the neural network always satisfies all currently known hard constraints. We therefore use BCP as a lightweight symbolic reasoning layer that filters illegal actions without requiring a full SAT solve after every decision.

\subsection{RL environment and reward}
\label{subsec:env_and_rew}

We use the same particle physics evaluation pipeline as AMBer~\cite{baretz2025towards} for direct comparison. A terminal state corresponds to a fully specified model matrix, which is passed to the physics tools. Using \texttt{Model2Mass}, the particle assignment matrix is compiled into a Lagrangian and the resulting Dirac and Majorana mass matrices are constructed, from which the light-neutrino mass matrix is obtained through the seesaw mechanism. Using \texttt{flavorpy}, the model's free parameters are then fit to neutrino observables from NuFit~5.3~\cite{Esteban:2020cvm} and scored by a chi-squared goodness of fit $\chi^2$. A model is \emph{invalid} if it fails a structural check (a rank-deficient mass matrix, so no physical spectrum exists) or is inelegant (more than $50$ free parameters); such models receive a strong penalty. Valid models are scored by a reward that combines (i) the chi-squared fit quality and (ii) the number of free parameters $n_p$, normalized and clipped to a bounded range so that better fits with fewer parameters score higher (an equal weighting of the two terms by default). We designate a valid model a \emph{neutrino model} when it additionally satisfies $\chi^2 \leq 10$ and $n_p \leq 7$. Because each evaluation requires expensive numerical optimization and dominates runtime (approx.\ 98\%; Section~\ref{sec:results}), avoiding invalid and redundant regions of the search space before evaluation is critical. Full experimental and compute details are given in Appendix~\ref{app:exp_details}.

\subsection{Certificate Extraction and Conflict Clauses}

Beyond scalar rewards, CDRL extracts structured certificates from invalid candidate models and converts them into reusable \textit{conflict clauses} added to the global constraint set $\mathcal{K}$. We implement this with a dedicated certificate-analysis routine that runs on every terminal model. It inspects the completed model matrix, tests it against the criteria below, and, when a criterion fires, identifies the exact subset of assignments responsible and emits their negation as a clause. Because a clause names the responsible assignments rather than the single candidate, one certificate prunes an entire class of models and lets the agent avoid revisiting similar configurations.
 We consider two main classes: (i) \emph{insufficient-rank representation certificates}, which arise when a candidate's non-Abelian representations and VEV choice yield a rank-deficient mass matrix for every choice of Abelian charge, and eliminate that specific representation-and-VEV combination independent of the charges; and (ii) \emph{$\mathbb{Z}_N$ symmetry certificates}, which depend only on the Abelian charges and capture two group-theoretic equivalences: charge assignments reducible to a smaller symmetry group by a shared divisor $d$ (checked in both directions between the smaller and larger group), and charge assignments related by negation $q \mapsto -q \bmod N$, an automorphism of order 2, both of which preserve the theory's selection rule. 
Table~\ref{tab:cert_types} demonstrates how each type of certificate works on a common $\mathbb{Z}_6$ candidate. Together, these certificates encode reusable symbolic knowledge that progressively removes invalid and redundant regions from the search space. 

\begin{table}[t]
\centering
\footnotesize
\setlength\tabcolsep{4pt}
\renewcommand{\arraystretch}{1.3}
\caption{\textbf{The three certificate types}, each worked through on a common $\mathbb{Z}_6$ candidate; the full derivation follows below. Certificates 2 and 3 depend only on the $\mathbb{Z}_N$ charges, independent of representations or VEVs. Every clause is $\neg(\text{the equivalent or excluded pattern})$; once added to the shared constraint database, it is enforced across all parallel searches with no physics evaluation spent on models it already rules out.}
\vspace{0.5cm}
\begin{tabular}{p{4.0cm} @{\hspace{0.4cm}}p{4.0cm} @{\hspace{0.4cm}}p{3.5cm}@{}}
\toprule
Description & Example & Clause \\
\midrule
\textbf{Insufficient-rank representation} \\ Model already rank-deficient without $\mathbb{Z}_N$ symmetry
& Reps and VEVs with $\mathbb{Z}_6$ charges wildcarded
already rank-deficient
& $\neg$(reps \& VEV, any charges) \\
\textbf{Abelian symmetry reducibility} \\ $\mathbb{Z}_N[q_i] \cong \mathbb{Z}_{N/d}[q_i/d]$ & $\mathbb{Z}_6\,[2,4,6] \mapsto \mathbb{Z}_3\,[1,2,3]$
& $\neg$(other representative) \\
\textbf{Charge negation} \\ $\mathbb{Z}_N[q_i] \cong \mathbb{Z}_{N}[-q_i \text{ mod N}]$
& $\mathbb{Z}_6\,[2,4,6] \mapsto  \mathbb{Z}_6\,[4,2,6]$
& $\neg$(negated charges) \\
\bottomrule
\end{tabular}
\label{tab:cert_types}
\end{table}

We now give the full construction of each certificate type. The first class arises when a candidate model's non-Abelian representations and VEV choice cannot support a full-rank mass matrix under any Abelian charge assignment. After constructing a candidate matrix, the system evaluates the model using the physics validity checker. If the model fails, the algorithm performs a reduced analysis in which the Abelian charge sector is temporarily wildcarded and only the non-Abelian representation and VEV assignments are retained. This has the effect of removing the Abelian symmetry entirely. When the representation-and-VEV-only model is already rank-deficient, no Abelian charge assignment layered on top of it can restore full rank, so the failure must be attributed solely to the non-Abelian representations and VEV choice. The system extracts the corresponding representation and VEV assignments and generates a conflict clause blocking this combination; other VEV choices for the same representations remain unaffected and available for exploration. This allows the solver to prune large classes of models that share the same insufficient-rank structure.

The second class arises from symmetry relations in the Abelian $\mathbb{Z}_N$ charge sector, independent of the non-Abelian representations or VEV choice, and captures two group-theoretic equivalences between charge assignments that describe the same physics. The first is reducibility. If all charges in a $\mathbb{Z}_N$ model share a common divisor $d$, then the model is reducible to a $\mathbb{Z}_{N/d}$ model. Formally, if
\[
q_i = d a_i, \qquad N = dM,
\]
then the selection rule
\[
q_1 + q_2 + q_3 = 0 \pmod{N}
\]
is equivalent to
\[
a_1 + a_2 + a_3 = 0 \pmod{M}.
\]
This implies that the larger symmetry provides no additional structure beyond the reduced group $\mathbb{Z}_M$. Whichever model is evaluated first, the certificate blocks the other, and the relation is checked in both directions (a smaller model also blocks the larger models it embeds into). The system generates a conflict clause by reconstructing the equivalent model in matrix form and negating the literals corresponding to that assignment. The second equivalence is relation by an automorphism, specifically the order-2 automorphism of negation. Replacing every charge with its additive inverse, $q \mapsto -q \bmod N$, preserves the selection rule $\sum_i q_i \equiv 0 \pmod N$ (negating every term preserves a sum of zero), so the negated charge assignment describes the same physics and need not be explored separately. Higher order automorphisms could also be included in future applications.

\subsection{Learned Clause Injection}

The conflict clauses generated during evaluation are accumulated globally and injected into the SAT solver before each new search episode, requiring no recompilation since each clause is already in CNF. Detection then falls out of BCP itself: at every MCTS node the current partial assignment is fed to the solver, and to score a candidate block assignment we tentatively add its literals and propagate. If propagation drives any learned clause to a conflict, that block assignment is removed from the legal-action mask before the policy network ever ranks it (Section~\ref{sec:sat}). Instead, a forced literal implied by the clauses is propagated into the state. Figure~\ref{fig:cdrl_arch} traces this loop end to end: a conflict at a completed model is analyzed into a clause, appended to the shared database, and the clause then masks the same failing pattern the moment it recurs elsewhere in the tree, before any physics evaluation is spent on it. This is precisely where the neuro-symbolic design pays off over a purely neural agent: the learned clause makes the whole equivalence class logically unreachable for every worker (Figure~\ref{fig:feedback_complementarity}), so large regions of the search space are pruned permanently rather than merely down-weighted.

\subsection{MCTS, Neural Guidance, and Portfolio Parallelization}

We search the space of partial model matrices using MCTS~\cite{silver2016mastering,Silver2017}. Each node corresponds to a partially constructed candidate. At each step, a policy-value network proposes the next assignment and estimates the state's value. The policy is masked by BCP's legal-action vector (Section~\ref{sec:sat}), ensuring only constraint-consistent actions are explored, and the masked priors guide tree traversal via PUCT~\cite{rosin2011multi}. At terminal states, the physics evaluation pipeline computes a reward that is backpropagated through the tree and used to retrain the network via self-play. This tightly couples all three components at every decision: neural guidance proposes, symbolic reasoning prunes, and MCTS balances exploration against exploitation. Removing any one causes substantial degradation and neutrino discovery collapses to near zero without either the network or MCTS (Table~\ref{tab:ablation}).

To scale search, we employ portfolio-style parallelization inspired by parallel SAT solvers~\cite{balyo2015hordesat,menouer2017parallel} and distributed RL~\cite{horgan2018distributed}. Multiple workers independently perform MCTS self-play while sharing (i) a replay buffer for network retraining and (ii) a global conflict clause database. Any certificate discovered by one worker is immediately enforced via BCP in all others, a direct analogue of clause sharing in portfolio SAT solving. Terminal evaluations are also cached globally to avoid redundant physics computation. Removing this sharing increases variance and reduces neutrino discovery by up to $2\times$ (Table~\ref{tab:ablation}). Additional details are in Appendices~\ref{app:expts:mctsnn} and~\ref{app:expts:parallel}.

\begin{table*}[!t]
\caption{\textbf{Comparison with Baselines across Theory Spaces.}
Valid (\%) denotes structurally valid models, Neutrino (\%) denotes discovered neutrino flavor models, and Min Params denotes the minimum number of free parameters found (lower is better). DNN: deep neural network (the policy-value network, Section~\ref{sec:sat}); PPO: Proximal Policy Optimization, the RL algorithm AMBer uses in place of CDRL's MCTS. Parenthetical multipliers under CDRL's Valid (\%) and Neutrino (\%) entries show improvement relative to AMBer, the prior state of the art; all other baselines (Random, ChatGPT, and the DNN/MCTS/symbolic ablations below AMBer) are shown for context and are not the basis for these multipliers. Each ablated or hybrid baseline (MCTS only, DNN only, Symbolic only, AlphaZero) is CDRL itself with the corresponding component removed, evaluated on the same physics pipeline described in Appendix~\ref{app:baseline}; this isolates that component's contribution rather than testing an independent reimplementation.}
\label{tab:main_comparison}
\setlength\tabcolsep{3.5pt}
\centering
\renewcommand{\arraystretch}{1.3}
\begin{adjustbox}{max width=1\textwidth}
\begin{tabular}{@{}l p{3.0cm} ccc ccc ccc@{}}
\toprule
\multirow{2}{*}{Method} 
& \multirow{2}{*}{Category}
& \multicolumn{3}{c}{$A_4 \times \mathbb{Z}_4$} 
& \multicolumn{3}{c}{$A_4 \times \mathbb{Z}_N$} 
& \multicolumn{3}{c}{$T_{19} \times \mathbb{Z}_4$} \\
\cmidrule(lr){3-5} \cmidrule(lr){6-8} \cmidrule(lr){9-11}
& 
& Valid (\%) & Neutrino (\%) & Min Params
& Valid (\%) & Neutrino (\%) & Min Params
& Valid (\%) & Neutrino (\%) & Min Params \\
\midrule

\multicolumn{11}{@{}l}{\textit{Uninformed and LLM baselines}} \\
Random 
& Random search 
& 0.1 & 0.00 & 12 
& 0.1 & 0.00 & 14 
& 0.1 & 0.00 & 16 \\

ChatGPT (GPT-5.3)~\cite{singh2025openai}
& Few-shot prompting 
& 0.2 & 0.00 & 13 
& 0.2 & 0.00 & 18 
& 0.1 & 0.00 & 15 \\

\hdashline

\multicolumn{11}{@{}l}{\textit{Search and learning hybrids}} \\
AlphaZero~\cite{silver2016mastering} 
& DNN + MCTS 
& 14.9 & 0.01 & 7 
& 16.2 & 0.03 & 7 
& 5.4 & 0.04 & 5 \\

MCTS only 
& Tree search only 
& 10.2 & 0.00 & 9 
& 9.8 & 0.00 & 10 
& 1.1 & 0.00 & 7 \\

DNN only 
& Policy/value network 
& 5.1 & 0.00 & 13 
& 4.5 & 0.00 & 9 
& 2.2 & 0.01 & 7 \\

Symbolic only~\cite{liang2016learning}  
& CDCL SAT solver
& 0.2 & 0.00 & 12 
& 0.2 & 0.00 & 12 
& 0.1 & 0.00 & 10 \\

\hdashline

\multicolumn{11}{@{}l}{\textit{State-of-the-art}} \\

AMBer~\cite{baretz2025towards}
& DNN + PPO 
& 19.53 & 0.02 & \textbf{5} 
& 13.79 & 0.03 & \textbf{5} 
& 11.02 & 0.03 & \textbf{4} \\

\midrule

\textbf{CDRL (Ours)} 
& \textbf{DNN + MCTS + certificates} 
& \makecell{\textbf{28.1 $\pm$ 3.3}\\{\scriptsize(1.44$\times$)}} 
& \makecell{\textbf{0.10 $\pm$ 0.01}\\{\scriptsize(5.00$\times$)}} 
& \textbf{5} 
& \makecell{\textbf{26.9 $\pm$ 3.7}\\{\scriptsize(1.95$\times$)}} 
& \makecell{\textbf{0.19 $\pm$ 0.05}\\{\scriptsize(6.33$\times$)}} 
& \textbf{5} 
& \makecell{\textbf{18.2 $\pm$ 2.1}\\{\scriptsize(1.65$\times$)}} 
& \makecell{\textbf{0.09 $\pm$ 0.02}\\{\scriptsize(3.00$\times$)}} 
& \textbf{4} \\

\bottomrule
\end{tabular}
\end{adjustbox}
\end{table*}

\subsection{Benchmark Theory Spaces}

We evaluate CDRL on neutrino flavor model discovery using the same theory spaces as AMBer~\cite{baretz2025towards} to ensure direct comparability. The search space consists of discrete flavor symmetry models defined by non-Abelian representations, Abelian $\mathbb{Z}_N$ charge assignments, and flavon VEV configurations. We consider three theory spaces used in prior work, defined by their symmetry structure and the number of flavons $n_\phi$:
(i) $A_4 \times \mathbb{Z}_4$ with $n_\phi \leq 5$ flavons,
(ii) $A_4 \times \mathbb{Z}_N$ where $N \in [2,10]$ and $n_\phi \leq 5$, and
(iii) $T_{19} \times \mathbb{Z}_4$ with $n_\phi \leq 6$. These spaces increase in symmetry richness and combinatorial complexity, providing progressively harder benchmarks.

\section{Experimental Results}
\label{sec:results}

\subsection{Evaluation Metrics}

We adopt the same evaluation protocol as AMBer to ensure consistency. Each candidate model is evaluated using the physics pipeline, which produces a $\chi^2$ fit to experimental observables and the number of free parameters $n_p$. A model is considered valid if it satisfies structural constraints, and is considered a neutrino model if it achieves both a good fit and low complexity, defined as $\chi^2 \leq 10$ and $n_p \leq 7$. We report the percentage of valid physics models (Valid \%), the percentage of viable neutrino models (Neutrino \%), and the minimum number of parameters discovered. 

\subsection{Main Results}

\begin{figure}[t]
\centering
\begin{minipage}{0.45\textwidth}
\centering
\hspace{-1.7cm}{\small\textbf{E2.Z4}}\\[0.5em]
\begin{tikzpicture}[scale=0.5, transform shape]\textbf{E2.Z4\_charge}
\node[cond] (r) {E1.Z4 = 3?};
\node[leaf, below left=1.6cm and 1cm of r] (a) {3};
\node[cond, below right=1.6cm and 1cm of r] (b) {E1.Z4 = 1?};
\node[leaf, below left=1.4cm and 1cm of b] (c) {1};
\node[leaf, below right=1.4cm and 1cm of b] (d) {2};
\draw[edge] (r) -- node[left] {\tiny Yes} (a);
\draw[edge] (r) -- node[right] {\tiny No} (b);
\draw[edge] (b) -- node[left] {\tiny Yes} (c);
\draw[edge] (b) -- node[right] {\tiny No} (d);
\end{tikzpicture}
\end{minipage}
\begin{minipage}{0.45\textwidth}
\centering
\hspace{-1.7cm}{\small\textbf{Hu.A4}}\\[0.5em]
\begin{tikzpicture}[scale=0.5, transform shape]
    \node[cond] (r) {E3.rep = 1''?};
    \node[leaf, below left=1.6cm and 1cm of r] (a) {1'};
    \node[cond, below right=1.6cm and 1cm of r] (b) {E3.rep = 1?};
    \node[leaf, below left=1.4cm and 1cm of b] (c) {1'};
    \node[leaf, below right=1.4cm and 1cm of b] (d) {1''};
    \draw[edge] (r) -- node[left] {\tiny Yes} (a);
    \draw[edge] (r) -- node[right] {\tiny No} (b);
    \draw[edge] (b) -- node[left] {\tiny Yes} (c);
    \draw[edge] (b) -- node[right] {\tiny No} (d);
    \end{tikzpicture}
    \end{minipage}
    \caption{\textbf{Example decision trees.} \textit{Left:} Determination of the $E_2$'s $\mathbb{Z}_4$ charge in $A_4\times\mathbb{Z}_4$. Among high-confidence states, this decision tree predicts the $E_2$'s $\mathbb{Z}_4$ charge from $E_1$'s $\mathbb{Z}_4$ charge. This demonstrates how CDRL has learned to match the $\mathbb{Z}_4$ charges of the charged lepton singlets, a reasonable result in order to construct physically realizable mass matrices. This observation also shows how some rules exclude possible choices (in this case the $\mathbb{Z}_4$ charge value of 4). \textit{Right:} Determination of the $H_u$'s $A_4$ representation in $A_4\times\mathbb{Z}_N$. This decision tree is harder to interpret physically. The up-flavor Higgs is not directly correlated with the charged lepton sector, so there is either some non-trivial relationship, or this is a spurious correlation within the CDRL search. The full set of such observations for all three theory spaces is in Appendix~\ref{app:kd_rules}.}
  \label{fig:kd_rule_example}
\end{figure}
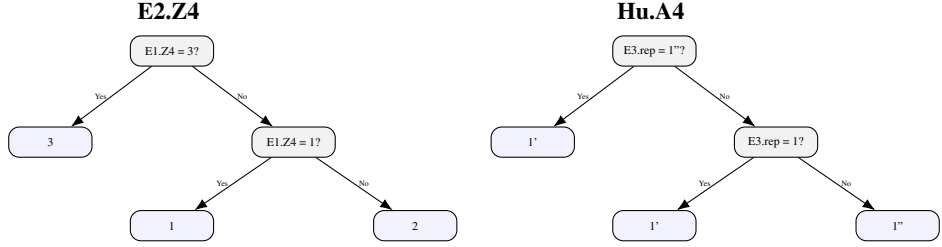

CDRL achieves the highest Valid (\%) and Neutrino (\%) of any method in every theory space, matching on AMBer's Min Params.  
 As shown in Table~\ref{tab:main_comparison}, on $A_4 \times \mathbb{Z}_N$, CDRL achieves $26.9\%$ valid models compared to $13.79\%$ for AMBer (a $1.95\times$ improvement), and $0.19\%$ neutrino models compared to $0.03\%$ (a $6.33\times$ improvement). 
 Similar gains are observed in other theory spaces, including up to $1.65\times$ improvements in valid model rates and up to $3\times$ improvements in neutrino discovery. These gains are achieved with substantially higher sample efficiency. As shown in Table~\ref{tab:search_efficiency}, CDRL discovers more neutrino models while evaluating up to $4\times$ fewer candidates. In combinatorial spaces exceeding $10^{26}$ possible models, these reductions correspond to substantially more efficient exploration. The mechanism behind these gains is concrete: Figure~\ref{fig:cdrl_arch} traces how a single certificate blocks an entire class of candidates before any physics is run, and Figure~\ref{fig:kd_rule_example} shows an example rule recovered from the search itself (full analysis in Section~\ref{sec:kd_results}).

\begin{table}[t]
\caption{\textbf{Search Efficiency Comparison.} CDRL discovers more neutrino models while evaluating significantly fewer candidates (up to $4\times$ fewer) across all theory spaces, demonstrating substantially improved sample efficiency over AMBer and random search.}
\label{tab:search_efficiency}
\setlength\tabcolsep{4pt}
\centering
\renewcommand{\arraystretch}{1}
\begin{tabular}{@{}llcc@{}}
\toprule
Space & Method & Models evaluated & Neutrino models found \\
\midrule

\multirow{3}{*}{$A_4 \times \mathbb{Z}_4$}
& Random & 4,000,000 & 28 \\
& AMBer & 4,000,000 & 683 \\
& CDRL  & 1,000,000 & \textbf{1090} \\

\noalign{\vskip 0.2ex}\hdashline\noalign{\vskip 0.4ex}

\multirow{3}{*}{$A_4 \times \mathbb{Z}_N$}
& Random & 4,000,000 & 21 \\
& AMBer & 4,000,000 & 1394 \\
& CDRL  & 1,000,000 & \textbf{2343} \\

\noalign{\vskip 0.2ex}\hdashline\noalign{\vskip 0.4ex}

\multirow{3}{*}{$T_{19} \times \mathbb{Z}_4$}
& Random & 24,000,000 & 409 \\
& AMBer & 24,000,000 & 6439 \\
& CDRL  & 6,000,000 & \textbf{7019} \\

\bottomrule
\end{tabular}
\end{table}

\paragraph{Importance of certificate-driven pruning.} The additional baselines (implementation details in Appendix~\ref{app:baseline}) are CDRL itself with one component removed, and highlight the importance of combining learning, planning, and symbolic reasoning. Few-shot prompting with GPT-5.3 performs near-random search, indicating that current large language models offer little advantage in these underexplored theory spaces: the relevant structure is largely absent from pretraining data, so the model cannot substitute for guided search with symbolic feedback. Pure symbolic search, DNN-only policies, MCTS-only search, and AlphaZero-style baselines all substantially underperform CDRL. On $A_4 \times \mathbb{Z}_N$ the AlphaZero-style baseline (MCTS and the policy-value network, without certificates) already exceeds AMBer's valid model rate ($16.2\%$ vs.\ $13.79\%$), so planning and neural guidance alone recover part of AMBer's gap to CDRL. Certificate-driven pruning (Table~\ref{tab:ablation}) accounts for the remainder and for most of the gain in neutrino discovery. Future work may study the additional performance metrics for such rule-guided search, beyond raw number of neutrino models found.

\begin{figure}[t]
    \centering
    \includegraphics[width=\textwidth]{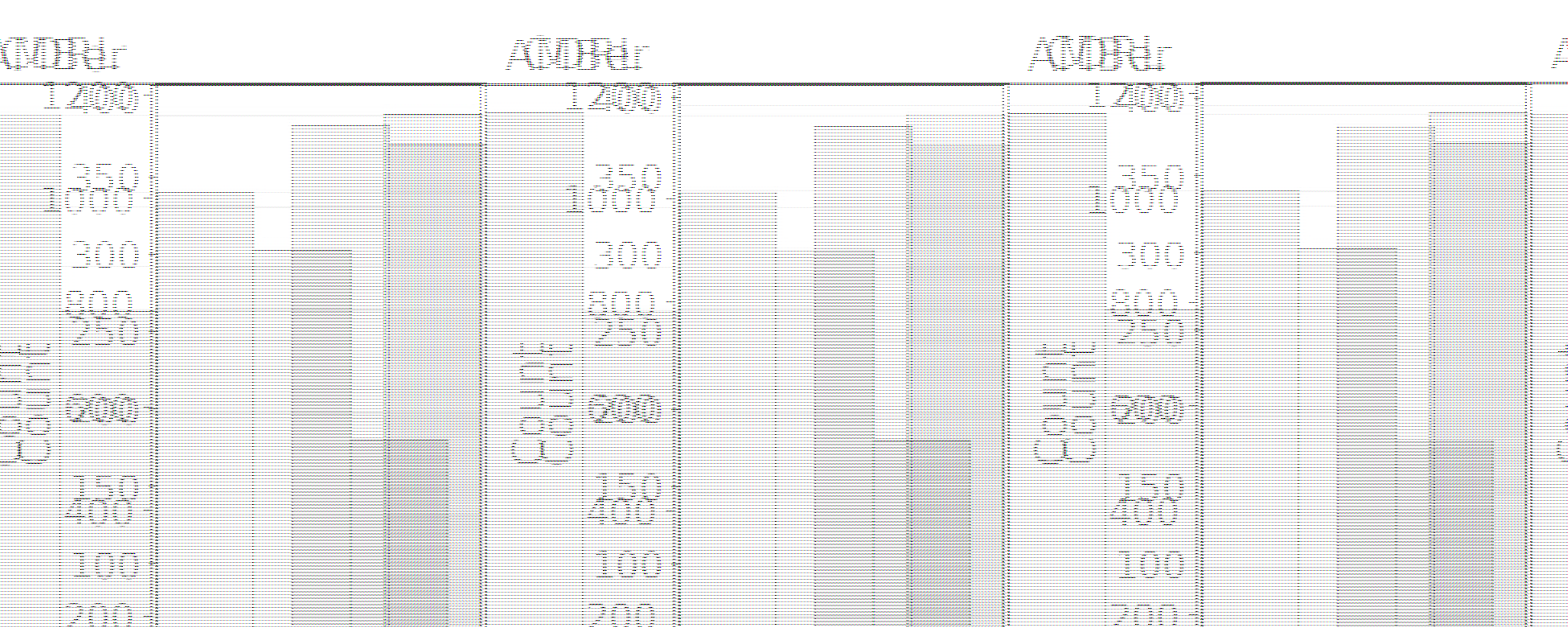}
    \caption{Distribution of successfully discovered neutrino models across $\mathbb{Z}_N$ symmetry groups for AMBer and CDRL. CDRL discovers viable models more uniformly across different symmetry choices, while AMBer concentrates more heavily around specific values of $N$.}
    \label{fig:zn}
\end{figure}

\begin{figure}[t]
    \centering
    \includegraphics[width=\textwidth]{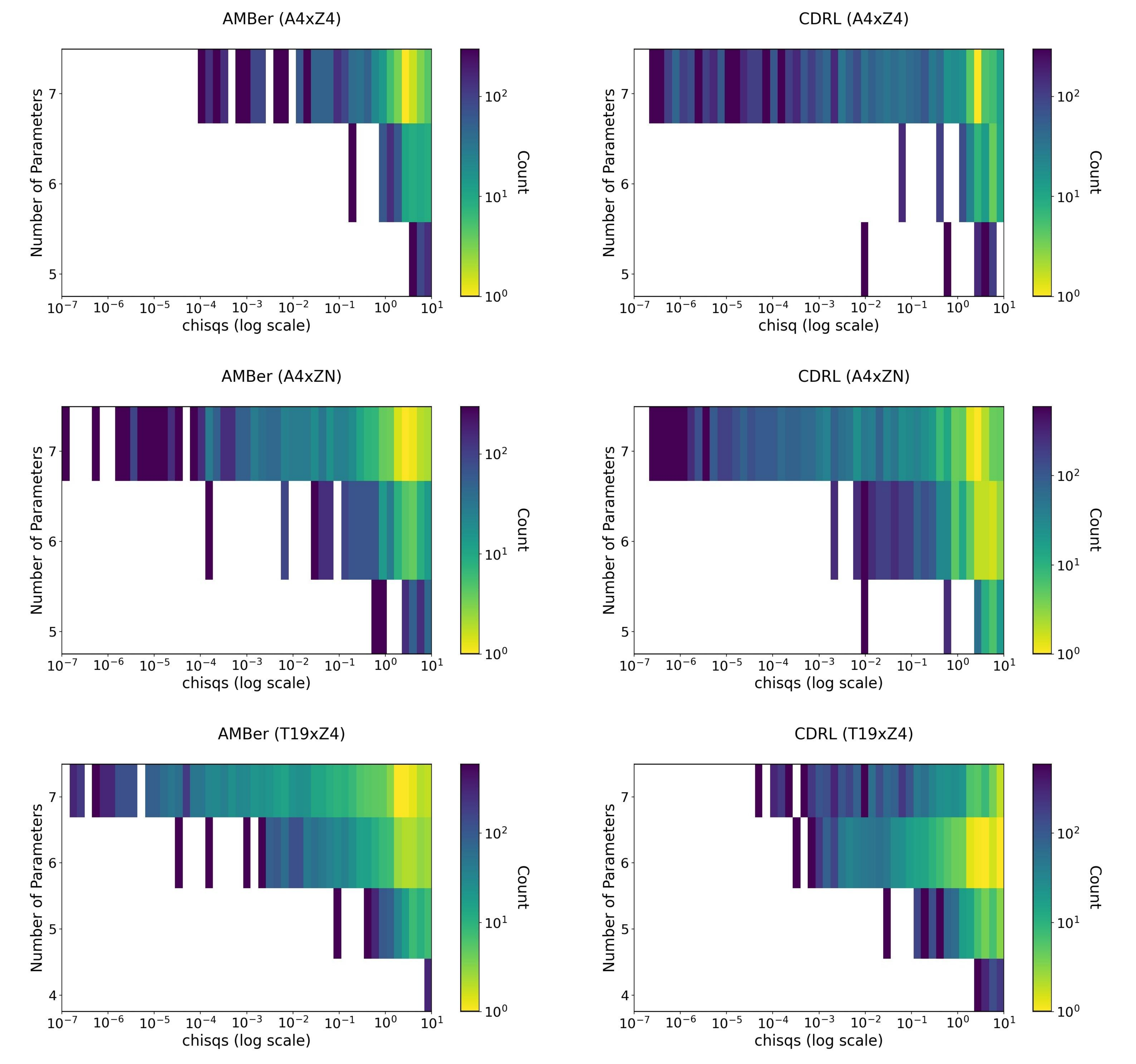}
    \caption{
Distribution of discovered models over $\chi^2$ (fit quality) and $n_p$ (number of parameters). Each bin shows model density, with the bottom-left region corresponding to well-fitting, low-complexity models. CDRL places more mass in the low-$n_p$ areas. Within that region, $\chi^2$ differences at very low values do not indicate meaningfully different fit quality, so Table~\ref{tab:main_comparison} and Table~\ref{tab:ablation} remain the primary quantitative comparison, with this figure illustrating where in $(\chi^2, n_p)$ space that advantage is concentrated.}
    \label{fig:heatplot}
\end{figure}

\paragraph{Qualitative trends.} 
Aggregate rates measure how often a search succeeds, but for model building the \emph{diversity} of what is found matters just as much: physicists want a broad set of distinct candidate theories to study, not many copies of the same one, and coverage across different symmetry groups indicates that the search is not confined to a narrow corner of the space. We therefore also examine the distribution of discovered models, not only their counts. Figure~\ref{fig:zn} shows the clearest evidence of this: CDRL discovers viable models spread across a substantially broader range of $\mathbb{Z}_N$ symmetry groups than AMBer, which concentrates around a few specific values of $N$. This indicates that CDRL explores different regions of the theory space rather than only different instances within the same region. Models are binned by the $\mathbb{Z}_N$ they were constructed under rather than any reduced or canonical form, so this spread is not an artifact of the same underlying model being counted under multiple symmetry groups; the reducibility certificate (Table~\ref{tab:cert_types}) instead prevents a redundant representative from being explored and counted a second time under a different $N$. Figure~\ref{fig:heatplot} shows where CDRL's models concentrate in $(\chi^2, n_p)$ space in the right column compared to AMBer in the left column. We also present a strong example model discovered by CDRL in the $T_{19} \times \mathbb{Z}_4$ space (Table~\ref{tab:t19z4_model}). The neutrino models discovered by CDRL do not exactly match any of those found by AMBer. Combined with the broader $\mathbb{Z}_N$ coverage in Figure~\ref{fig:zn}, this is consistent with CDRL exploring complementary regions of the theory space rather than merely finding different individual models within the same region AMBer already covers.

Following the analysis protocol of AMBer~\cite{baretz2025towards}, we can look more closely at why these distributions differ. For the distribution of models shown in Figure~\ref{fig:heatplot}, CDRL reaches neutrino models with the smallest parameter counts observed for either method (Table~\ref{tab:main_comparison} reports this as Min Params) while maintaining competitive or lower $\chi^2$ values. For the $\mathbb{Z}_N$ distribution of Figure~\ref{fig:zn}, AMBer concentrates its discoveries around specific symmetry groups (e.g., $N=5$), whereas CDRL spreads more uniformly across the available $\mathbb{Z}_N$ choices. This trend is consistent with progressively stronger symbolic pruning during search. As the learned clause database grows, the legal-action mask becomes increasingly selective, allowing Boolean constraint propagation to eliminate larger regions of invalid or redundant search trajectories before expensive physics evaluation is performed. This mechanism both concentrates evaluation on predictive low-parameter models and, because whole redundant equivalence classes are removed rather than merely down-weighted, pushes the search into $\mathbb{Z}_N$ regions AMBer rarely reaches. CDRL's reward contains no term that rewards higher-order $\mathbb{Z}_N$ symmetries (Section~\ref{subsec:env_and_rew} gives the full reward, which depends only on $\chi^2$ fit quality and the number of free parameters); AMBer's scalar reward includes such a term to steer search toward the more complex, higher-order groups~\cite{baretz2025towards}. CDRL reaches broader coverage without an equivalent incentive. Certificates prune redundant equivalence classes uniformly across the theory space, so exploration spreads more evenly as a byproduct of pruning rather than of reward shaping. This difference in reward design does not affect the evaluation protocol itself (Section~\ref{sec:results}), which scores both methods identically.

Although CDRL introduces slight additional overhead from MCTS, certificate analysis, and symbolic propagation, the dominant computational cost (approx.\ 98\%) arises from the physics evaluation pipeline, which is the same as AMBer. Since experiments use different parallelization settings and compute configurations (CDRL uses 32 CPU cores, while AMBer uses 250 parallel environments), we report candidate evaluations as the primary normalized measure of search efficiency.

\subsection{Ablation Study}

An ablation study measures each component's importance by removing it from the full system and observing the resulting drop in performance, so a larger drop indicates a more important component. Table~\ref{tab:ablation} isolates the contribution of each component this way. Error bars are the standard deviation across 4 runs with different random seeds. Every ablation degrades performance, confirming that the gains come from integrating learning, planning, and symbolic reasoning rather than from any single part. Removing either the neural network or MCTS collapses neutrino discovery to near zero, showing that guided exploration is necessary at all in a space this large. Removing symbolic certificates instead leaves a functioning search, with valid model rates falling from $26.9\%$ to $16.2\%$ and neutrino discovery from $0.19\%$ to $0.03\%$ on $A_4 \times \mathbb{Z}_N$, so certificates contribute a substantial improvement on top of the neural and planning components. Removing portfolio sharing lowers performance and raises variance, indicating that shared constraints and experience are critical for scaling. In short, each component contributes measurably, and the full system is required for strong performance.

\begin{table*}[!t]
\caption{\textbf{Ablation Study of CDRL Components.}
Valid (\%) denotes the percentage of structurally valid models, Neutrino (\%) denotes the percentage of discovered neutrino flavor models, and Min Params denotes the minimum number of free parameters found.}
\label{tab:ablation}
\setlength\tabcolsep{3.5pt}
\centering
\renewcommand{\arraystretch}{1.3}
\begin{adjustbox}{max width=1\textwidth}
\begin{tabular}{@{}l >{\raggedright\arraybackslash}p{3.0cm} ccc ccc ccc@{}}
\toprule
\multirow{2}{*}{Method / Ablation} 
& \multirow{2}{*}{Description}
& \multicolumn{3}{c}{$A_4 \times \mathbb{Z}_4$} 
& \multicolumn{3}{c}{$A_4 \times \mathbb{Z}_N$} 
& \multicolumn{3}{c}{$T_{19} \times \mathbb{Z}_4$} \\
\cmidrule(lr){3-5} \cmidrule(lr){6-8} \cmidrule(lr){9-11}
& 
& Valid (\%) & Neutrino (\%) & Min Params
& Valid (\%) & Neutrino (\%) & Min Params
& Valid (\%) & Neutrino (\%) & Min Params \\
\midrule

\textbf{CDRL (Full)} 
& DNN + MCTS + symbolic certificates + portfolio 
& \textbf{28.1 $\pm$ 3.3} & \textbf{0.10 $\pm$ 0.01} & \textbf{5} 
& \textbf{26.9 $\pm$ 3.7} & \textbf{0.19 $\pm$ 0.05} & \textbf{5} 
& \textbf{18.2 $\pm$ 2.1} & \textbf{0.09 $\pm$ 0.02} & \textbf{4} \\

\midrule

No DNN 
& No policy/value network 
& 10.2 $\pm$ 5.1 & 0.00 $\pm$ 0.00 & 9 
& 9.8 $\pm$ 4.6 & 0.00 $\pm$ 0.00 & 10 
& 1.1 $\pm$ 1.0 & 0.00 $\pm$ 0.00 & 7 \\

No MCTS 
& No planning 
& 5.1 $\pm$ 8.4 & 0.00 $\pm$ 0.00 & 13 
& 4.5 $\pm$ 9.0 & 0.00 $\pm$ 0.00 & 9 
& 2.2 $\pm$ 3.6 & 0.01 $\pm$ 0.00 & 7 \\

No symbolic 
& No certificate 
& 14.9 $\pm$ 3.1 & 0.01 $\pm$ 0.02 & 7 
& 16.2 $\pm$ 2.1 & 0.03 $\pm$ 0.01 & 7 
& 5.4 $\pm$ 4.9 & 0.04 $\pm$ 0.02 & 5 \\

No portfolio 
& No clause/replay shared
& 26.9 $\pm$ 6.5 & 0.04 $\pm$ 0.02 & 6 
& 25.1 $\pm$ 7.2 & 0.10 $\pm$ 0.09 & 5 
& 12.1 $\pm$ 7.8 & 0.06 $\pm$ 0.04 & 5 \\

\bottomrule
\end{tabular}
\end{adjustbox}
\end{table*}

\subsection{Search Knowledge Discovery}

\label{sec:kd_results}

\begin{figure}[htbp]
    \centering
    \includegraphics[width=\textwidth]{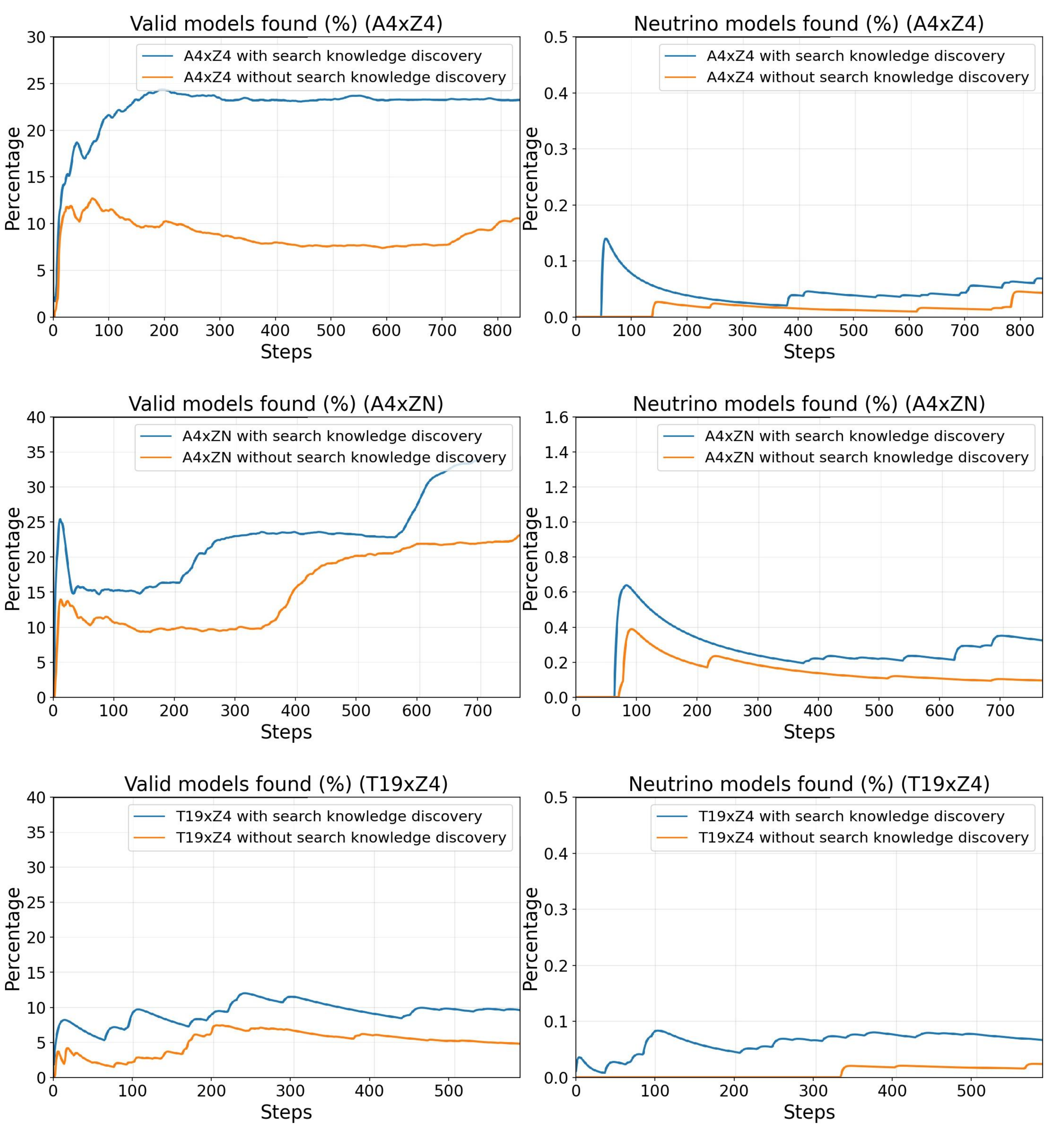}
    \caption{Impact of search knowledge discovery (KD) across theory spaces. Each panel is titled by metric and theory space. We compare CDRL with and without rule-based soft constraints on $A_4 \times \mathbb{Z}_4$, $A_4 \times \mathbb{Z}_N$, and $T_{19} \times \mathbb{Z}_4$. Incorporating rules extracted from high-confidence search trajectories consistently improves both valid model rates and neutrino model discovery. Rule-guided runs achieve higher final performance and faster convergence, with gains of roughly $1.5$--$2\times$ in valid models and $2$--$3\times$ in neutrino models across the three spaces, demonstrating that the extracted rules provide useful and generalizable guidance for exploration. These experiments are performed using smaller-scale KD evaluation runs, so absolute values are not directly comparable to the full-scale results reported in Table~\ref{tab:main_comparison}.}
    \label{fig:kd}
\end{figure}

A by-product of the search is a record of the decisions CDRL makes on its way to good models. We examine this record post-hoc to recover common decision patterns from the MCTS trajectories and cast them into human-readable decision trees. These decision patterns both illuminate how the agent is exploring, and with sufficient interpretation by a human physicist can provide insight on common modes of viable models in the theory space.
Our approach is related to prior work on extracting interpretable policies from RL systems~\cite{bastani2018verifiable}. However, in addition to using extracted rules purely for explanation or verification, we inject these rules as soft symbolic constraints in fresh, smaller-scale, ``rule-guided'' CDRL runs to test their utility in deepening exploration of fruitful regions of the theory space. Effectively, these soft constraints encourage the agent towards known viable regions of the theory space while still permitting exploration of the entire space. As shown in Figure~\ref{fig:kd}, incorporating these rules consistently increases the number of discovered models in all three theory spaces. In $A_4 \times \mathbb{Z}_4$ and $T_{19} \times \mathbb{Z}_4$, rule-guided runs achieve $\sim\!2\times$ higher valid model rates and $2$--$3\times$ higher neutrino model discovery, while in $A_4 \times \mathbb{Z}_N$ we observe $\sim\!1.5\times$ improvements in validity and up to $3\times$ in neutrino discovery.

We describe the procedure used to extract soft constraints from MCTS trajectories in Algorithm~\ref{alg:soft-constraints} and incorporate them into the search process. We retain high-confidence decision points from MCTS trajectories, \textit{i.e.} states where the dominant action receives more than $95\%$ of visits and has a positive value estimate, and group these states by construction progress, defined as the number of assigned entries in the state before the first unassigned position. For each progress bucket with enough samples, we train one shallow decision tree that predicts the dominant action from the partial model's earlier assignments; hyperparameters are given in Appendix~\ref{app:exp_details}. Because blocks are assigned in a fixed top-down order (Section~\ref{sec:application}), only blocks assigned earlier in that order are available as features. We extract these high-confidence root-to-leaf paths as rules of the form $\text{conditions} \rightarrow a$, where the conditions are conjunctions over binary state features, and retain only rules expressible as binary conditions for compatibility with the symbolic representation. These rules can then be injected into a new CDRL run as soft constraints.
At inference time, we identify all rules whose conditions are satisfied by the current state, let each activated rule vote for its preferred action to form a bias vector $r(s)$, and reweight the neural policy as $\tilde{P}(a \mid s) \propto P(a \mid s)\,(1 + \beta\, r_a(s))$, where $\beta = 0.5$ in all experiments. The reweighted policy is then masked by the legality constraints and renormalized before being used in MCTS, so soft constraints guide the search without overriding hard constraints or eliminating exploration.

\begin{algorithm}[t]
\caption{Soft Constraint Extraction and Policy Reweighting}
\label{alg:soft-constraints}
\begin{algorithmic}[1]

\Require MCTS traces $\mathcal{D}$, progress buckets $\mathcal{B}$
\Require Minimum samples $m$, tree hyperparameters $(d, s, \ell)$
\Require Soft bias strength $\beta$

\State Collect state-action samples $(x_i, a_i)$ from MCTS
\State Retain high-confidence samples based on visit concentration and value estimates
\State Group samples into buckets $\{\mathcal{D}_b\}_{b \in \mathcal{B}}$ based on progress

\For{each bucket $b \in \mathcal{B}$}
    \If{$|\mathcal{D}_b| < m$}
        \State continue
    \EndIf

    \State Train decision tree $T_b$ with depth $d$, split $s$, leaf size $\ell$
    \State Extract leaf paths corresponding to high-confidence predictions
    \State Convert each path into a rule $(\text{conditions} \rightarrow a)$
    \State Store rules with support and purity
\EndFor

\State Build rule table indexed by progress and decision block

\Statex

\Statex \textbf{Policy Reweighting during MCTS:}

\State Given state $s$, compute neural policy $P(\cdot \mid s)$
\State Identify rules satisfied by $s$ and compute bias vector $r(s)$
\State Reweight policy:
\[
\tilde{P}(a \mid s) \propto P(a \mid s)\,(1 + \beta\, r_a(s))
\]
\State Apply legality mask and renormalize $\tilde{P}$

\end{algorithmic}
\end{algorithm}

The extracted rules are simple if-else conditions over particle assignments that reveal recurring structure in successful models. For example, assignments of charged lepton singlets often follow consistent patterns (e.g., $E_1$ influencing $E_2, E_3$ through identical or systematically shifted $\mathbb{Z}_4$ charges). We also observe coordinated triplet assignments across $L$, $N$, and $\phi$, mirroring common design patterns used by human model builders~\cite{Ding:2011gt}. Although extracted post-hoc, these rules provide reusable and generalizable structure. The full set of 40 rules is listed in Section~\ref{app:kd_rules}.

Crucially, interpreting each rule within the model building process requires additional context not always captured within the decision tree, and many rules are more representative of the agent's search process than of correlations between parameters of the model. For example, several of the observed rules for the $\mathbb{Z}_N$ charges of the charged lepton singlets ($E_1, E_2, E_3$) show them favoring identical or closely related charges. This is a reasonable pattern for correctly producing their masses if other particles have already been given appropriate representation assignments. However, some rules restrict themselves to only a subset of the charge values the symmetry allows. This demonstrates the limitation of interpreting these rules physically.
Each tree is fit only to the high-confidence trajectories within a single progress bucket, a small, selective sample. Therefore, a tree can look like a high-confidence rule simply because the values outside it were rare in that particular sample, not because the wider search avoids them. 
While we introduce a tool for search knowledge discovery, further work in this direction may enable more robust interpretation of these discovered correlations in terms of useful model-building rules.
Surfacing this kind of observation is still useful. It either flags a rule that will not hold up once tested on more data, or, when a pattern holds up under closer inspection, a regularity of the theory space a human model builder had not already anticipated.

\section{Conclusion, Limitation, and Future Work}
\label{sec:conclusions}

We introduced CDRL, a framework for scientific discovery in large constrained hypothesis spaces. Unlike standard RL methods that rely primarily on scalar rewards, CDRL uses structured certificates from symbolic domain-specific reasoning tools to identify why candidates fail and convert these failures into reusable constraints. On neutrino flavor model discovery, CDRL consistently improves search efficiency across all theory spaces. Per candidate evaluated, it achieves up to 1.95$\times$ higher valid model rates and 6.33$\times$ higher neutrino model discovery than AMBer. Separately, reaching a comparable number of neutrino models requires up to 4$\times$ fewer physics evaluations (Table~\ref{tab:search_efficiency}). CDRL also improves coverage of the theory space, matches AMBer's best model complexity while discovering far more of them, and enables post-hoc search knowledge discovery through interpretable rules extracted from search trajectories.

The core idea of CDRL is not specific to neutrino physics, nor to particle physics. Many scientific domains pair a large discrete design space with an expensive verifier and an external tool that can explain failures, including chemistry, materials science, and automated theorem proving. In each, the dominant cost is the verification step, so the same computational bottleneck we address here applies: a scalar reward wastes that expensive budget by letting the agent revisit equivalent failures, whereas a certificate removes the whole failing class once and for all. Wherever a reasoning tool can name the assignments responsible for a violation, CDRL can turn that explanation into a constraint that is reused across the search. Finally, the rules recovered through search knowledge discovery are useful beyond efficiency, as they compress what the agent learned into human-readable structure, which both accelerates future runs and gives scientists a first look at the regularities of a mathematical space they are only beginning to explore. We have not yet independently validated these rules as physically meaningful. At this stage, they describe regularities in how the search itself explores the space, which are themselves shaped by our design choices, and confirming which of them reflect genuine physical structure is left to future work with domain experts.

The main limitation of CDRL is that it depends on the availability of useful certificate-generating tools and is currently best suited to structured, discrete search spaces where constraints can be represented and reused effectively. The framework could also introduce overhead in larger search spaces from certificate analysis and clause management. Future work will extend CDRL to broader scientific domains such as chemistry and materials science, where external verifiers can provide rich feedback. Another promising direction is to learn when and how to generalize certificates into reusable constraints, and to combine CDRL with foundation models so that neural priors, symbolic pruning, and scientific verification can work together in a unified discovery loop.

\section*{Acknowledgments}
PJ was supported by a seed grant from the AI4Science Center in the College of Sciences at Georgia Tech.  MF is supported by Fermilab which is administered by
Fermi Forward Discovery Group, LLC under Contract
No. 89243024CSC000002 with the U.S. Department of
Energy, Office of Science, Office of High Energy Physics. JR is supported by a fellowship from the WATCHEP training grant from the U.S. Department of Energy, Office of Science. The work of VKP was supported in part by the U.S. National Science Foundation under Grant PHY-221028.

\newpage
{
\small
\bibliographystyle{ieeetr}
\bibliography{main}
}


\appendix

\begin{figure}[h]
    \centering
    \textbf{The CDRL loop, step by step} \\[0.2em]
    {\small\color{gray}one MCTS tree, shown as it is built}\\[0.6em]
    \includegraphics[width=0.5\textwidth]{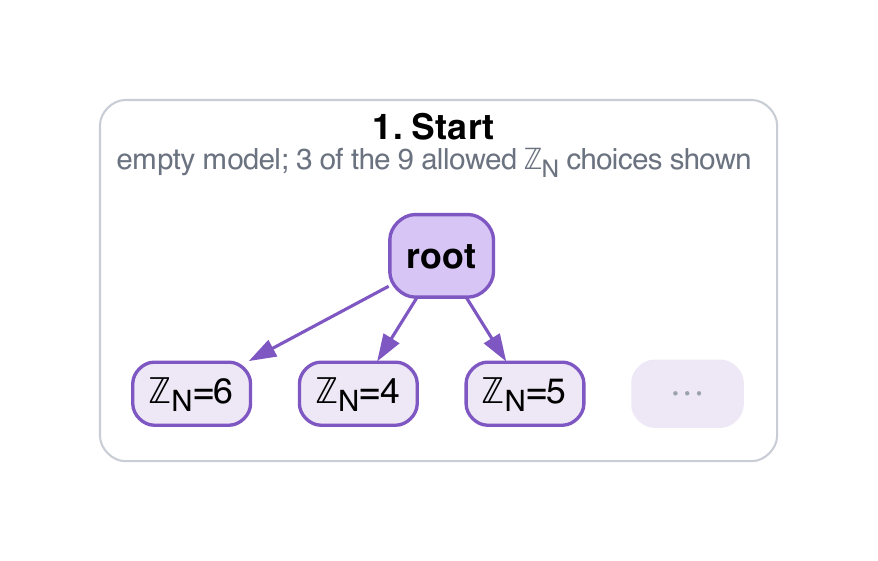}\\[0.6em]
    \includegraphics[width=0.5\textwidth]{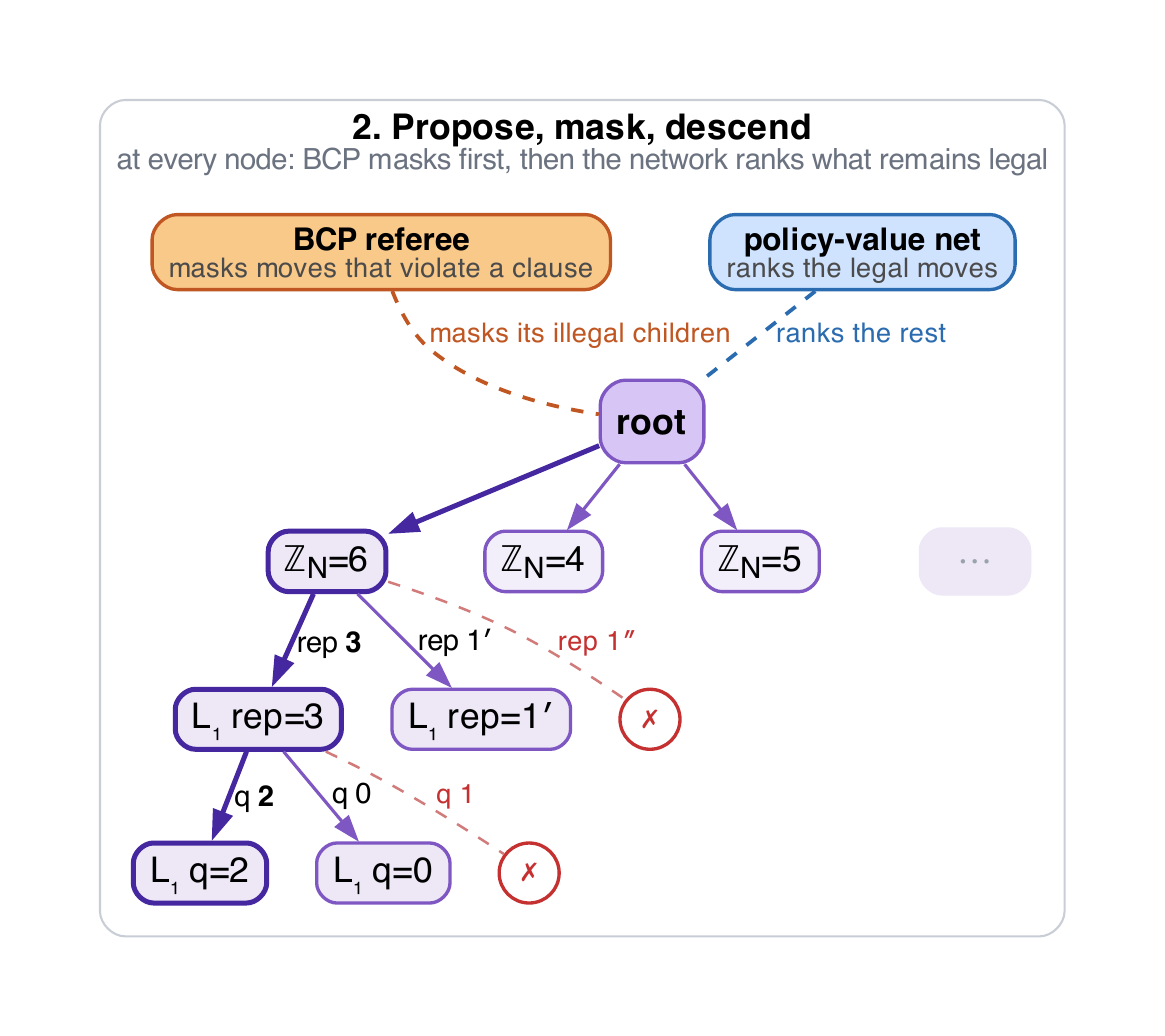}
    \caption{\textbf{The CDRL mechanism, shown as four snapshots of the same tree.} \textbf{(1) Start:} the empty model; three of the nine allowed $\mathbb{Z}_N$ choices are shown at the root. \textbf{(2) Propose, mask, descend:} at the root, and again at every node MCTS visits, the BCP referee masks the illegal children first and the policy-value network then ranks what remains; MCTS commits to $\mathbb{Z}_N=6$ and descends, and the same red $\times$ masking recurs at each node along the way. Continued below.}
    \label{fig:mechanism_tree_steps}
\end{figure}

\begin{figure}[h]
    \ContinuedFloat
    \centering
    \includegraphics[width=0.5\textwidth]{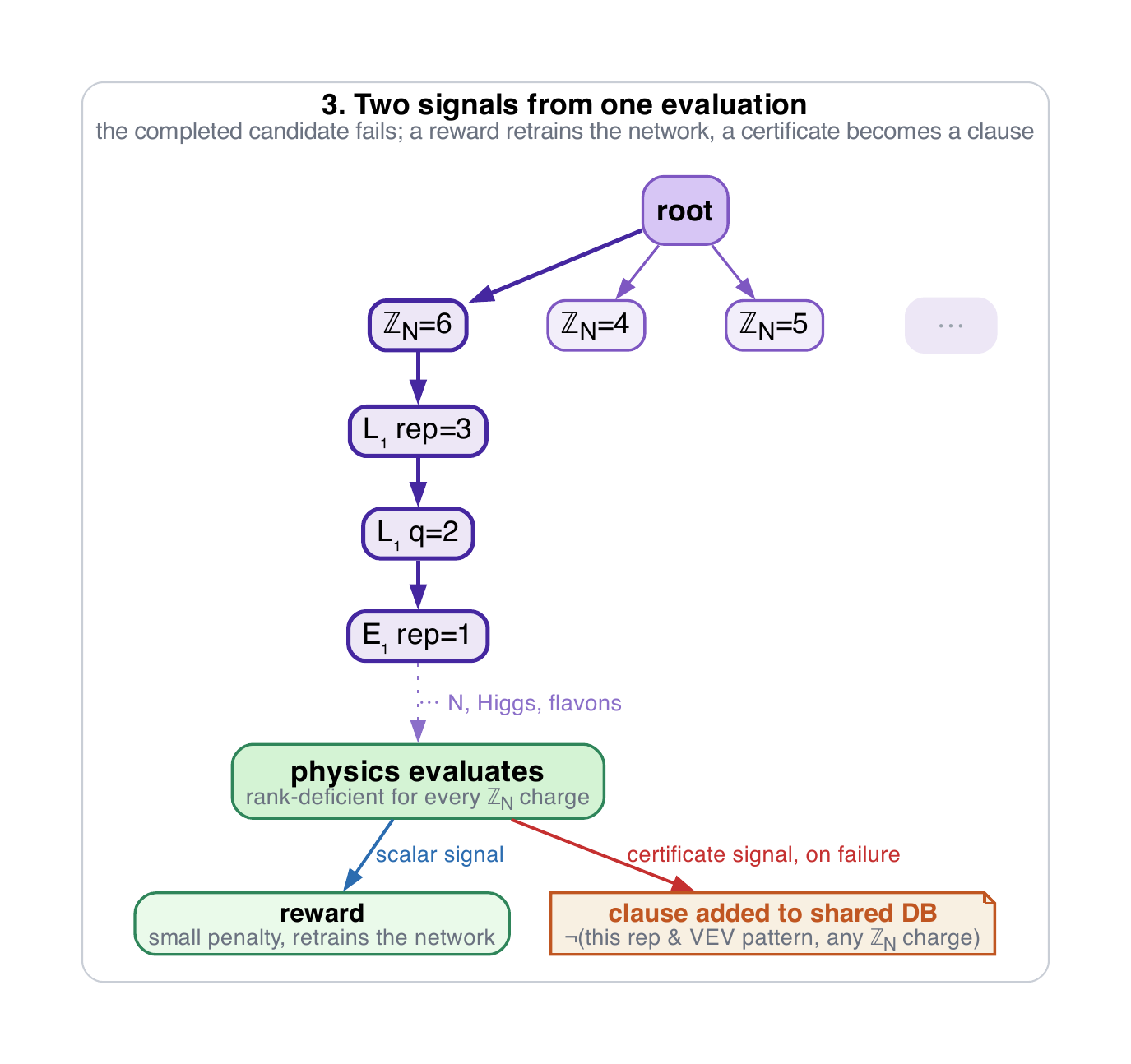}\\[0.6em]
    \includegraphics[width=0.5\textwidth]{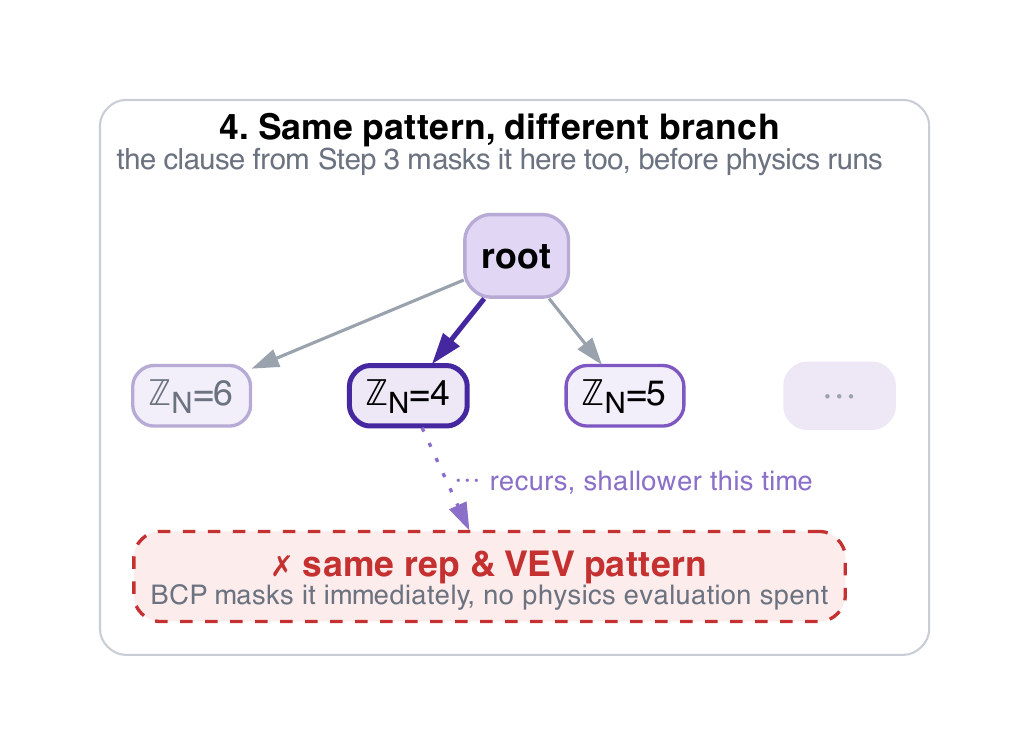}
    \caption{\textbf{(continued)} \textbf{(3) Two signals from one evaluation:} the branch completes and the candidate fails, rank-deficient for every $\mathbb{Z}_N$ charge; this single evaluation produces a scalar reward that retrains the network and, separately, a certificate that the analyzer converts into a clause added to the shared database. \textbf{(4) Same pattern, different branch:} MCTS next explores the sibling $\mathbb{Z}_N=4$ branch, and the clause learned in step 3 masks the same representation-and-VEV pattern as soon as it recurs, before the branch is built out further or any physics evaluation is spent on it.}
\end{figure}

\section{The CDRL Loop: A Step-by-Step Walkthrough}
\label{app:mechanism_steps}

Figure~\ref{fig:mechanism_tree_steps} below walks through a worked example as four snapshots of the same tree, taken as it is actually built, covering the same components as Figure~\ref{fig:cdrl_arch} (the policy-value network, the BCP referee, MCTS, physics evaluation, the certificate analyzer, and the clause database) except search knowledge discovery, which operates offline after training and is not part of this per-candidate loop. Showing the same tree across panels also makes the key claim concrete rather than only described: a clause learned from one branch permanently prunes a matching pattern anywhere else in the tree.

At every node, not only the root, the BCP referee first masks any child that would violate an already-learned clause, and the policy-value network then ranks the remaining legal children; this repeats at each subsequent decision (the $\mathbb{Z}_N$ choice, then each particle's representation, then its charge, and so on). What Figure~\ref{fig:mechanism_tree_steps} shows as a single selected path is itself the outcome of many such simulations from the current state, each one descending through already-expanded nodes by upper-confidence bound and expanding one new leaf; the move actually played is the one visited most often across these simulations, not simply the network's top-ranked action. As in Algorithm~\ref{alg:cdrl}, the network is updated periodically from replay-buffer mini-batches rather than immediately after every single evaluation, and the certificate path in Step 3 only triggers when the completed candidate fails; a valid candidate produces a reward alone. The three $\mathbb{Z}_N$ choices at the root are an illustrative subset of the nine allowed in this theory space ($N \in \{2,\dots,10\}$).

\section{Additional Experimental Details}
\label{app:exp_details}

\subsection{Compute Infrastructure}
\label{subsec:compute}

All experiments are conducted on a machine with Intel Xeon Gold 6448Y CPUs (32 cores, 2.10 GHz), 32\,GB RAM, and an NVIDIA H100 GPU (80GB, MIG 20GB partition). Parallel self-play is executed across CPU workers, while neural network training is performed on the GPU.

The dominant computational cost in our setting arises from the physics evaluation pipeline, which involves symbolic model construction, numerical optimization, and parameter fitting for candidate theories. MCTS traversal and SAT-based propagation introduce additional overhead, but are comparatively lightweight relative to terminal model evaluation. Since different methods use different parallelization settings and compute configurations (CDRL uses 32 CPU cores, while AMBer uses 250 parallel environments), we report candidate evaluations as the primary normalized efficiency metric.

\subsection{Additional Implementation Details}

Our framework is implemented in PyTorch. We integrate symbolic constraint reasoning using a SAT solver backend, leveraging MiniSAT-style Boolean Constraint Propagation (BCP) via the PySAT library~\cite{pysat}. 

Domain constraints are compiled into conjunctive normal form (CNF) and enforced during search. After each action, the solver applies BCP to propagate implied assignments, ensuring that all intermediate states remain consistent with hard constraints.

For search knowledge discovery rule extraction (Section~\ref{sec:kd_results}), we treat an MCTS decision as high-confidence when the dominant action receives more than $95\%$ of visits and has a positive value estimate. Progress buckets require at least $10$ such samples; each decision tree has maximum depth $3$, requires at least $10$ samples to split a node and $5$ samples per leaf, and is fit with an 80/20 train-test split when enough data is available.

\subsection{Baselines}
\label{app:baseline}

We compare against \textbf{AMBer} (Autonomous Model Builder)~\cite{baretz2025towards}, an RL framework for neutrino flavor model discovery. AMBer performs iterative model refinement using scalar rewards derived from the physics evaluation pipeline, including $\chi^2$ fit quality, the number of free parameters, and a bonus for exploring higher-order $\mathbb{Z}_N$ symmetry groups. Unlike CDRL, AMBer does not use certificate-driven symbolic pruning or reusable constraint learning.

We additionally compare against several ablated and hybrid baselines, including pure MCTS search, DNN-only policies, symbolic-only SAT-based search~\cite{liang2016learning}, and a DNN+MCTS framework~\cite{silver2016mastering}. These baselines isolate the contributions of planning, neural guidance, and symbolic reasoning.

\subsubsection{LLM Baseline Details}

We evaluate a prompting-based baseline using GPT-5.3~\cite{singh2025openai}. The model is prompted to directly generate candidate neutrino flavor model matrices satisfying the required representation, charge, and VEV assignment format used throughout the paper.

The prompt contains: (i) a task description explaining the flavor model construction problem, (ii) formatting instructions specifying the matrix structure and allowed assignments, and (iii) 3 few-shot examples of valid candidate model matrices drawn from the corresponding theory space. The model is then asked to generate new candidate matrices autoregressively.

Generated outputs are parsed into matrix form and evaluated using the same symbolic constraints and physics evaluation pipeline used for all other methods. Invalid, incomplete, or unparsable outputs are counted as failed candidates.

We use temperature 0.8 and top-p 0.95 decoding for generation. The baseline does not use search, SAT-based propagation, iterative refinement, or certificate-driven feedback, and therefore serves as a reference point for pure prompting-based generation without structured exploration.

\subsection{Neural Network Architecture}

The policy-value network operates on a flattened matrix representation of the state. The architecture consists of:

\begin{itemize}
\item Fully connected layer: 512 units with BatchNorm and ReLU
\item Fully connected layer: 256 units with BatchNorm and ReLU
\item Policy head: linear layer to action dimension followed by log-softmax
\item Value head: linear layer followed by $\tanh$
\end{itemize}

\subsection{Additional Training Details}
\label{app:expts:mctsnn}

We provide the concrete hyperparameters for the MCTS and network training described in Section~\ref{sec:application}. We use:

\begin{itemize}
\item Max iterations: 30
\item Episodes per iteration: 100
\item MCTS simulations per move: 150
\item MCTS $c_{\text{puct}}$: 0.5
\item Batch size: 512
\item Training epochs per iteration: 20
\item Optimizer: Adam
\end{itemize}

The policy head is trained using cross-entropy loss against MCTS visit counts, while the value head is trained using mean squared error against normalized terminal rewards.

After each training iteration, the updated model is evaluated against the previous checkpoint in a single-agent self-play evaluation setting. Both models independently generate candidate solutions using the same environment and evaluation pipeline, and we compare their cumulative terminal rewards across a fixed number of evaluation episodes. The checkpoint is updated only if the new model achieves higher cumulative reward than the previous model. Unlike standard AlphaZero pit-play, this comparison does not involve a competitive two-player game, since neutrino model discovery is a single-agent environment.

\subsection{Parallelization}
\label{app:expts:parallel}

We employ a portfolio-style parallelization strategy inspired by parallel SAT solvers. Multiple workers independently perform MCTS-guided self-play while sharing:

\begin{itemize}
\item neural network replay buffers
\item a global database of learned conflict clauses
\end{itemize}

Conflict clauses discovered by any worker are synchronized globally and reused by all workers. Additionally, expensive terminal evaluations are cached to avoid redundant computation.

This combination of parallel search and clause sharing significantly improves exploration efficiency and sample efficiency.

In our current implementation, learned clauses are accumulated monotonically and shared across workers, and we do not employ CDCL-style clause deletion. In practice, this remains computationally manageable because SAT propagation constitutes only a small fraction of the total runtime, while the dominant cost arises from the physics evaluation pipeline. Although the number of learned certificates grows over training, we did not observe clause management becoming a bottleneck in the theory spaces considered in this work. For larger domains, clause forgetting or activity-based clause management would be a natural extension.

\section{Search Knowledge Discovery Rule Galleries}
\label{app:kd_rules}

Below we list the full set of observations extracted from the high-confidence MCTS trajectories in each theory space, using the procedure described in Section~\ref{sec:kd_results}. Each observation is a shallow decision tree that predicts one block assignment (a representation or a $\mathbb{Z}_N$ charge) from earlier assignments in the same partial model; an observation that collapses to a single leaf is a \emph{constant assignment}, always taking the same value in the high-confidence set. For every tree we report the train and test accuracy of the predicted assignment. In the two-way trees, the left branch is the \emph{Yes} outcome of the condition and the right branch is \emph{No}. These rules are extracted post-hoc and reused as the soft policy bias of Algorithm~\ref{alg:soft-constraints}; the aggregate effect of reusing them is reported in Figure~\ref{fig:kd}.

\subsection{$A_4 \times \mathbb{Z}_4$}

\begin{center}\textbf{Constant assignments}\end{center}
\begin{center}
\begin{tikzpicture}
\node[kdnode] (a) {L1.rep = 3};
\node[kdnode, right=2.5cm of a] (b) {E1.rep = 1''};
\node[kdnode, right=2.5cm of b] (c) {E1.Z4 = 4};
\node[kdnode, below=1.2cm of a] (d) {E2.rep = 1};
\node[kdnode, right=2.5cm of d] (e) {E3.rep = 1'};
\node[kdnode, right=2.5cm of e] (f) {N1.rep = 3};
\node[kdnode, below=1.2cm of d] (g) {phi1.rep = 3};
\node[kdnode, right=2.5cm of g] (h) {phi2.rep = 3};
\end{tikzpicture}
\end{center}

\begin{center}\textbf{E2.Z4\_charge}\quad{\small Train: 0.917 \quad Test: 0.917}\end{center}
\begin{center}
\begin{tikzpicture}[scale=0.9, transform shape]
\node[cond] (r) {E1.Z4 = 3?};
\node[leaf, below left=1.6cm and 2cm of r] (a) {3};
\node[cond, below right=1.6cm and 2cm of r] (b) {E1.Z4 = 1?};
\node[leaf, below left=1.4cm and 1.4cm of b] (c) {1};
\node[leaf, below right=1.4cm and 1.4cm of b] (d) {2};
\draw[edge] (r) -- node[left] {\tiny Yes} (a);
\draw[edge] (r) -- node[right] {\tiny No} (b);
\draw[edge] (b) -- node[left] {\tiny Yes} (c);
\draw[edge] (b) -- node[right] {\tiny No} (d);
\end{tikzpicture}
\end{center}

\begin{center}\textbf{E3.Z4\_charge}\quad{\small Train: 0.972 \quad Test: 0.944}\end{center}
\begin{center}
\begin{tikzpicture}[scale=0.9, transform shape]
\node[cond] (r) {E2.Z4 = 1?};
\node[leaf, below left=1.6cm and 2cm of r] (a) {1};
\node[cond, below right=1.6cm and 2cm of r] (b) {E1.Z4 = 4?};
\node[leaf, below left=1.4cm and 1.4cm of b] (c) {4};
\node[leaf, below right=1.4cm and 1.4cm of b] (d) {2};
\draw[edge] (r) -- node[left] {\tiny Yes} (a);
\draw[edge] (r) -- node[right] {\tiny No} (b);
\draw[edge] (b) -- node[left] {\tiny Yes} (c);
\draw[edge] (b) -- node[right] {\tiny No} (d);
\end{tikzpicture}
\end{center}

\begin{center}\textbf{N1.Z4\_charge}\quad{\small Train: 0.911 \quad Test: 1.000}\end{center}
\begin{center}
\begin{tikzpicture}[scale=0.9, transform shape]
\node[cond] (r) {L2.Z4 = 3?};
\node[leaf, below right=1.6cm and 2cm of r] (a) {2};
\node[cond, below left=1.6cm and 2cm of r] (b) {E3.Z4 = 1?};
\node[leaf, below left=1.4cm and 1.4cm of b] (c) {2};
\node[leaf, below right=1.4cm and 1.4cm of b] (d) {4};
\draw[edge] (r) -- node[right] {\tiny Yes} (a);
\draw[edge] (r) -- node[left] {\tiny No} (b);
\draw[edge] (b) -- node[left] {\tiny Yes} (c);
\draw[edge] (b) -- node[right] {\tiny No} (d);
\end{tikzpicture}
\end{center}

\begin{center}\textbf{Hd.rep}\quad{\small Train: 0.615 \quad Test: NA}\end{center}
\begin{center}
\begin{tikzpicture}
\node[cond] (r) {E3.rep = 1'?};
\node[leaf, below left=1.6cm and 2cm of r] (a) {1};
\node[leaf, below right=1.6cm and 2cm of r] (b) {1'};
\draw[edge] (r) -- node[left] {\tiny Yes} (a);
\draw[edge] (r) -- node[right] {\tiny No} (b);
\end{tikzpicture}
\end{center}

\begin{center}\textbf{phi1.Z4\_charge}\quad{\small Train: 0.654 \quad Test: 0.429}\end{center}
\begin{center}
\begin{tikzpicture}[scale=0.9, transform shape]
\node[cond] (r) {Hu.Z4 = 2?};
\node[leaf, below left=1.6cm and 2cm of r] (a) {3};
\node[cond, below right=1.6cm and 2cm of r] (b) {Hu.Z4 = 4?};
\node[leaf, below left=1.4cm and 1.4cm of b] (c) {3};
\node[cond, below right=1.4cm and 1.4cm of b] (d) {E1.Z4 = 1?};
\node[leaf, below left=1.4cm and 1.4cm of d] (e) {4};
\node[leaf, below right=1.4cm and 1.4cm of d] (f) {3};
\draw[edge] (r) -- node[left] {\tiny Yes} (a);
\draw[edge] (r) -- node[right] {\tiny No} (b);
\draw[edge] (b) -- node[left] {\tiny Yes} (c);
\draw[edge] (b) -- node[right] {\tiny No} (d);
\draw[edge] (d) -- node[left] {\tiny Yes} (e);
\draw[edge] (d) -- node[right] {\tiny No} (f);
\end{tikzpicture}
\end{center}

\begin{center}\textbf{phi2.Z4\_charge}\quad{\small Train: 0.719 \quad Test: 0.444}\end{center}
\begin{center}
\begin{tikzpicture}
\node[cond] (r) {Hu.Z4 = 1?};
\node[leaf, below left=1.6cm and 2cm of r] (a) {4};
\node[leaf, below right=1.6cm and 2cm of r] (b) {2};
\draw[edge] (r) -- node[left] {\tiny Yes} (a);
\draw[edge] (r) -- node[right] {\tiny No} (b);
\end{tikzpicture}
\end{center}

\subsection{$A_4 \times \mathbb{Z}_N$}

\begin{center}\textbf{Constant assignments}\end{center}
\begin{center}
\begin{tikzpicture}
\node[kdnode] (a) {L1.rep = 3};
\node[kdnode, right=2.5cm of a] (b) {E2.rep = 1'};
\node[kdnode, right=2.5cm of b] (c) {N1.rep = 3};
\node[kdnode, below=1.2cm of a] (d) {phi1.rep = 3};
\node[kdnode, right=2.5cm of d] (e) {phi2.rep = 3};
\node[kdnode, right=2.5cm of e] (f) {phi3.rep = 3};
\node[kdnode, below=1.2cm of d] (g) {phi4.rep = 3};
\end{tikzpicture}
\end{center}

\begin{center}\textbf{E2.Z5\_charge}\quad{\small Train: 0.938 \quad Test: 0.882}\end{center}
\begin{center}
\begin{tikzpicture}[scale=0.9, transform shape]
\node[cond] (r) {E1.Z5 = 2?};
\node[leaf, below left=1.6cm and 2cm of r] (a) {2};
\node[cond, below right=1.6cm and 2cm of r] (b) {E1.Z5 = 1?};
\node[leaf, below left=1.4cm and 1.4cm of b] (c) {1};
\node[leaf, below right=1.4cm and 1.4cm of b] (d) {4};
\draw[edge] (r) -- node[left] {\tiny Yes} (a);
\draw[edge] (r) -- node[right] {\tiny No} (b);
\draw[edge] (b) -- node[left] {\tiny Yes} (c);
\draw[edge] (b) -- node[right] {\tiny No} (d);
\end{tikzpicture}
\end{center}

\begin{center}\textbf{E3.rep}\quad{\small Train: 0.707 \quad Test: 0.727}\end{center}
\begin{center}
\begin{tikzpicture}
\node[cond] (r) {E2.rep = 1'?};
\node[leaf, below left=1.6cm and 2cm of r] (a) {1''};
\node[leaf, below right=1.6cm and 2cm of r] (b) {1};
\draw[edge] (r) -- node[left] {\tiny Yes} (a);
\draw[edge] (r) -- node[right] {\tiny No} (b);
\end{tikzpicture}
\end{center}

\begin{center}\textbf{E3.Z5\_charge}\quad{\small Train: 0.926 \quad Test: 0.968}\end{center}
\begin{center}
\begin{tikzpicture}[scale=0.9, transform shape]
\node[cond] (r) {E2.Z5 = 4?};
\node[leaf, below left=1.6cm and 2cm of r] (a) {4};
\node[cond, below right=1.6cm and 2cm of r] (b) {E1.Z5 = 2?};
\node[leaf, below left=1.4cm and 1.4cm of b] (c) {2};
\node[cond, below right=1.4cm and 1.4cm of b] (d) {E2.Z5 = 1?};
\node[leaf, below left=1.4cm and 1.4cm of d] (e) {1};
\node[leaf, below right=1.4cm and 1.4cm of d] (f) {5};
\draw[edge] (r) -- node[left] {\tiny Yes} (a);
\draw[edge] (r) -- node[right] {\tiny No} (b);
\draw[edge] (b) -- node[left] {\tiny Yes} (c);
\draw[edge] (b) -- node[right] {\tiny No} (d);
\draw[edge] (d) -- node[left] {\tiny Yes} (e);
\draw[edge] (d) -- node[right] {\tiny No} (f);
\end{tikzpicture}
\end{center}

\begin{center}\textbf{N1.Z5\_charge}\quad{\small Train: 0.861 \quad Test: 0.900}\end{center}
\begin{center}
\begin{tikzpicture}[scale=0.9, transform shape]
\node[cond] (r) {E2.Z5 = 3?};
\node[leaf, below right=1.6cm and 2cm of r] (a) {3};
\node[cond, below left=1.6cm and 2cm of r] (b) {E1.Z5 = 2?};
\node[leaf, below left=1.4cm and 1.4cm of b] (c) {2};
\node[cond, below right=1.4cm and 1.4cm of b] (d) {L3.Z5 = 5?};
\node[leaf, below left=1.4cm and 1.4cm of d] (e) {4};
\node[leaf, below right=1.4cm and 1.4cm of d] (f) {2};
\draw[edge] (r) -- node[right] {\tiny Yes} (a);
\draw[edge] (r) -- node[left] {\tiny No} (b);
\draw[edge] (b) -- node[left] {\tiny Yes} (c);
\draw[edge] (b) -- node[right] {\tiny No} (d);
\draw[edge] (d) -- node[left] {\tiny Yes} (e);
\draw[edge] (d) -- node[right] {\tiny No} (f);
\end{tikzpicture}
\end{center}

\begin{center}\textbf{Hu.rep}\quad{\small Train: 0.889 \quad Test: 0.600}\end{center}
\begin{center}
\begin{tikzpicture}
\node[cond] (r) {E3.rep = 1''?};
\node[leaf, below left=1.6cm and 2cm of r] (a) {1'};
\node[cond, below right=1.6cm and 2cm of r] (b) {E3.rep = 1?};
\node[leaf, below left=1.4cm and 1.4cm of b] (c) {1'};
\node[leaf, below right=1.4cm and 1.4cm of b] (d) {1''};
\draw[edge] (r) -- node[left] {\tiny Yes} (a);
\draw[edge] (r) -- node[right] {\tiny No} (b);
\draw[edge] (b) -- node[left] {\tiny Yes} (c);
\draw[edge] (b) -- node[right] {\tiny No} (d);
\end{tikzpicture}
\end{center}

\subsection{$T_{19} \times \mathbb{Z}_4$}
{\small Left edge: Yes \quad Right edge: No}

\begin{center}\textbf{L1.Z4\_charge}\quad{\small Train: 0.792 \quad Test: 0.714}\end{center}
\begin{center}\begin{tikzpicture}\node[leaf] {4};\end{tikzpicture}\end{center}

\begin{center}\textbf{E1.Z4\_charge}\quad{\small Train: 0.611 \quad Test: 0.600}\end{center}
\begin{center}
\begin{tikzpicture}
\node[cond] {L2.Z4 = 4?}
    child { node[leaf] {1} edge from parent }
    child { node[cond] {L1.Z4 = 1?} edge from parent
        child { node[leaf] {2} edge from parent }
        child { node[leaf] {1} edge from parent } };
\end{tikzpicture}
\end{center}

\begin{center}\textbf{E2.rep}\quad{\small Train: 0.950 \quad Test: 0.905}\end{center}
\begin{center}
\begin{tikzpicture}
\node[cond] {E1.rep = 1?}
    child { node[leaf] {1'} edge from parent }
    child { node[leaf] {1} edge from parent };
\end{tikzpicture}
\end{center}

\begin{center}\textbf{E2.Z4\_charge}\quad{\small Train: 0.938 \quad Test: 0.857}\end{center}
\begin{center}
\begin{tikzpicture}
\node[cond] {E1.Z4 = 2?}
    child { node[leaf] {2} edge from parent }
    child { node[cond] {E1.Z4 = 1?} edge from parent
        child { node[leaf] {1} edge from parent }
        child { node[leaf] {3} edge from parent } };
\end{tikzpicture}
\end{center}

\begin{center}\textbf{E3.rep}\quad{\small Train: 0.685 \quad Test: 0.694}\end{center}
\begin{center}
\begin{tikzpicture}
\node[cond] {E1.rep = 1''?}
    child { node[cond] {E2.rep = 1''?} edge from parent
        child { node[leaf] {1'} edge from parent }
        child { node[leaf] {1''} edge from parent } }
    child { node[leaf] {1} edge from parent };
\end{tikzpicture}
\end{center}

\begin{center}\textbf{E3.Z4\_charge}\quad{\small Train: 0.924 \quad Test: 0.926}\end{center}
\begin{center}
\begin{tikzpicture}
\node[cond] {E1.Z4 = 2?}
    child { node[leaf] {2} edge from parent }
    child { node[cond] {E2.Z4 = 4?} edge from parent
        child { node[leaf] {4} edge from parent }
        child { node[cond] {E2.Z4 = 3?} edge from parent
            child { node[leaf] {3} edge from parent }
            child { node[leaf] {1} edge from parent } } };
\end{tikzpicture}
\end{center}

\begin{center}\textbf{N1.rep}\quad{\small Train: 0.960 \quad Test: 0.932}\end{center}
\begin{center}
\begin{tikzpicture}
\node[cond] {L2.rep = 3\_3b?}
    child { node[leaf] {3\_3} edge from parent }
    child { node[leaf] {3\_3b} edge from parent };
\end{tikzpicture}
\end{center}

\begin{center}\textbf{Hu.rep}\quad{\small Train: 0.522 \quad Test: 0.667}\end{center}
\begin{center}
\begin{tikzpicture}
\node[cond] {E3.rep = 1?}
    child { node[leaf] {1''} edge from parent }
    child { node[cond] {N3.rep = 3\_2b?} edge from parent
        child { node[leaf] {1'} edge from parent }
        child { node[cond] {N1.rep = 3\_3?} edge from parent
            child { node[leaf] {1'} edge from parent }
            child { node[leaf] {1} edge from parent } } };
\end{tikzpicture}
\end{center}

\begin{center}\textbf{Hu.Z4\_charge}\quad{\small Train: 0.978 \quad Test: 0.833}\end{center}
\begin{center}\begin{tikzpicture}\node[leaf] {4};\end{tikzpicture}\end{center}

\begin{center}\textbf{Hd.rep}\quad{\small Train: 0.821 \quad Test: 0.875}\end{center}
\begin{center}
\begin{tikzpicture}
\node[cond] {N3.rep = 3\_2?}
    child { node[leaf] {1''} edge from parent }
    child { node[cond] {E2.rep = 1?} edge from parent
        child { node[leaf] {1''} edge from parent }
        child { node[leaf] {1} edge from parent } };
\end{tikzpicture}
\end{center}

\begin{center}\textbf{phi1.rep}\quad{\small Train: 0.859 \quad Test: 0.941}\end{center}
\begin{center}
\begin{tikzpicture}
\node[cond] {L3.rep = 3\_3b?}
    child { node[leaf] {3\_2b} edge from parent }
    child { node[cond] {E1.rep = 1?} edge from parent
        child { node[leaf] {3\_3} edge from parent }
        child { node[leaf] {3\_3b} edge from parent } };
\end{tikzpicture}
\end{center}

\begin{center}\textbf{phi2.rep}\quad{\small Train: 0.750 \quad Test: 0.625}\end{center}
\begin{center}
\begin{tikzpicture}
\node[cond] {N2.rep = 3\_3?}
    child { node[leaf] {3\_2b} edge from parent }
    child { node[cond] {E1.rep = 1''?} edge from parent
        child { node[leaf] {3\_2} edge from parent }
        child { node[cond] {N3.rep = 3\_2?} edge from parent
            child { node[leaf] {3\_1} edge from parent }
            child { node[leaf] {3\_3} edge from parent } } };
\end{tikzpicture}
\end{center}

\begin{center}\textbf{phi3.rep}\quad{\small Train: 0.741 \quad Test: 0.714}\end{center}
\begin{center}
\begin{tikzpicture}
\node[cond] {N1.rep = 3\_3?}
    child { node[leaf] {3\_2b} edge from parent }
    child { node[leaf] {3\_1} edge from parent };
\end{tikzpicture}
\end{center}

\begin{center}\textbf{phi5.rep}\quad{\small Train: 0.857 \quad Test: 0.750}\end{center}
\begin{center}
\begin{tikzpicture}
\node[leaf] {3\_2b};
\end{tikzpicture}
\end{center}

\end{document}